\documentclass[10pt,twocolumn,letterpaper]{article}

\usepackage{wacv}              

\usepackage{graphicx}
\usepackage{amsmath}
\usepackage{amssymb}
\usepackage{booktabs}
\usepackage{algorithm}
\usepackage{algpseudocode}
\usepackage{tabularx,multirow}
\usepackage{times}
\usepackage{epsfig}
\usepackage{nicefrac}
\usepackage[dvipsnames]{xcolor}
\usepackage{soul}
\usepackage[accsupp]{axessibility}  

\usepackage[pagebackref,breaklinks,colorlinks]{hyperref}
\usepackage{etoolbox}

\usepackage[capitalise]{cleveref}
\crefname{section}{Sec.}{Secs.}
\Crefname{section}{Section}{Sections}
\Crefname{table}{Table}{Tables}
\crefname{table}{Tab.}{Tabs.}

\def\wacvPaperID{713} 
\def\confName{WACV}
\def\confYear{2024}

\begin{document}

\title{GC-MVSNet: Multi-View, Multi-Scale, Geometrically-Consistent \\
  Multi-View Stereo}

\author{Vibhas K. Vats, Sripad Joshi, David J. Crandall\\
Indiana University Bloomington\\
{\tt\small \{vkvats, joshisri, djcran\}@iu.edu}
\and
Md. Alimoor Reza\\
Drake University\\
{\tt\small md.reza@drake.edu}
\and 
Soon-heung Jung\\
ETRI\\
{\tt\small zeroone@etri.re.kr}
}

\newcommand\myfigure{%
    \vspace{-25pt}
  \includegraphics[width=1\textwidth,height=12\baselineskip]{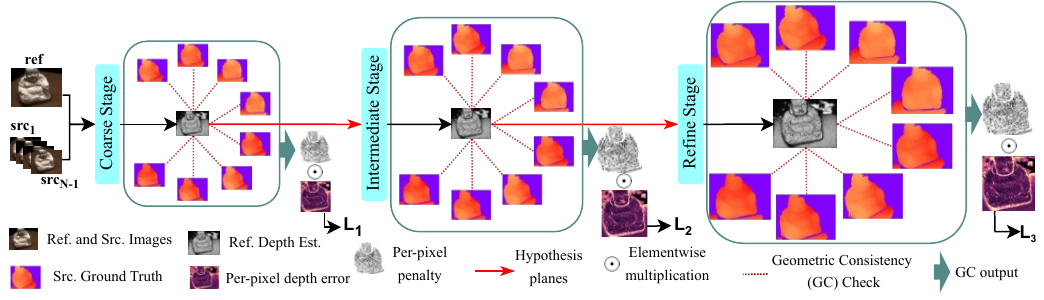}\\
  \refstepcounter{figure}\normalfont Figure~\thefigure: Our
  multi-view, multi-scale geometric consistency checking
  process. During training, the geometric consistency of the estimated
  depth map is explicitly modeled across multiple source views. This allows the model to more quickly and accurately learn about geometric consistency,
  allowing the trained model to produce better reconstructions during inference.
  \label{fig:gc-module-idea}\\
}

\makeatletter
\let\@oldmaketitle\@maketitle
\renewcommand{\@maketitle}{\@oldmaketitle
    \myfigure}
\makeatother

\maketitle



\begin{abstract}
\vspace{-10pt}
Traditional multi-view stereo (MVS) methods rely heavily on
photometric and geometric consistency constraints, but
newer machine learning-based MVS methods check geometric
consistency across multiple source views only as a
post-processing step. In this paper, we present a novel 
approach that explicitly encourages
geometric consistency of reference view depth maps across
multiple source views at different scales during learning (see \cref{fig:gc-module-idea}).
We find that adding this geometric consistency loss
significantly accelerates learning by explicitly
penalizing geometrically inconsistent pixels,  
reducing the training iteration requirements to nearly half
that of other MVS methods.
Our extensive experiments show that our approach 
achieves a new state-of-the-art on the DTU and
BlendedMVS datasets, and competitive results on the Tanks and
Temples benchmark. To the best of our knowledge,
GC-MVSNet is the first attempt to enforce multi-view, multi-scale 
geometric consistency during learning.
\end{abstract}


\vspace{-18pt}
\section{Introduction}\label{sec:intro}

Traditional multi-view stereo (MVS) methods such as Gipuma
\cite{Galliani2015fusibile}, Furu \cite{Furukawa2010AccurateDA},
COLMAP \cite{Johannes2016pixelwise}, and Tola
\cite{tola2011largescale} rely on solving for photometric and
geometric consistency constraints across multiple views.  Recent
machine learning-based MVS methods \cite{ding2022transmvsnet,
  peng2022rethinkingMVS, Luo2019Pmvsnet, yao2019recurrent,
  chen2019pointbased, gu2019casmvsnet, yang2019CVPMVS,
  Cheng2019USCNet, wei2021aa, xu2019multiscale, yao2018mvsnet,
  YU2021AAcostvolume} use deep networks to extract feature maps and
then construct 3D cost volumes to measure similarity between feature
maps \cite{gu2019casmvsnet}.
Each paper in this stream of machine learning-based approaches    has introduced innovations that have significantly improved
the quality of depth estimates and point cloud
reconstructions,
like multi-level feature 
extraction, attention-based feature matching, similarity-based 
cost-volume creation, and improved loss formulations.
%
%
%
These modern methods use plane sweep
volumes to implicitly encode geometric constraints,
and perform multi-view geometric
consistency checks as a postprocess after inference to filter the depth maps.
However,
they do not
explicitly 
model
multi-view geometric constraints during learning. Instead,
learning about multi-view geometric
information thus must happen only implicitly.

In this paper, we show, for the first time, that providing the model with explicit multi-view geometric
cues using geometric consistency checks across multiple source
views during training  (see Fig. \ref{fig:gc-module-idea}) significantly improves
accuracy while significantly lowering the training iteration requirements.
We formulate a multi-stage model called GC-MVSNet which learns
geometric cues at three scales.
At each scale,
we introduce a novel multi-view geometric consistency module that performs geometric
consistency checks of reference view depth estimates across multiple
source views and generates a per-pixel penalty. This
penalty is then combined with per-pixel depth error
(estimated using cross-entropy loss at each stage) to generate the final loss.

This formulation of loss function provides abundant geometric cues to
accelerate learning of the model. Our extensive experiments show that
GC-MVSNet requires nearly
half the training iterations needed by other recent models \cite{yao2018mvsnet,
  gu2019casmvsnet, peng2022rethinkingMVS, wei2021aa,
  ding2022transmvsnet}. Our approach also achieves a new state-of-the-art accuracy 
on DTU \cite{jensen2014dtu} and BlendedMVS \cite{yao2019blended}
datasets, and competitive results on
Tanks and Temples
\cite{Knapitsch2017tnt}. To the best
of our knowledge, GC-MVSNet is the first attempt to leverage
multi-view, multi-scale geometric consistency checks during the
training process. We also perform extensive ablation experiments to
demonstrate the effectiveness of the proposed approach.

In summary, in this paper:
\begin{itemize}
\setlength\itemsep{-0.1em}
    \item[--] We propose a novel multi-view, multi-scale geometric consistency (GC) module during learning that encourages geometric consistency of reference view depth maps across multiple source views.
    \item[--] We show that this technique reduces the training iteration requirements to
      nearly half that of other  models,
        by explicitly providing multi-view geometric cues during learning.
    \item[--] We show that the module is highly general and can be plugged into different MVS pipelines to enhance geometric cues during training.
\end{itemize}

\section{Related Work}\label{sec:related works}

The taxonomy proposed by Furukawa and Ponce
\cite{Furukawa2010AccurateDA} classifies MVS methods into four primary
scene representations: \textit{volumetric fields}
\cite{kutulakos1999spacecarving, seitz1997photorealistic,
  faugeras1998, sinha2007graphcut}, \textit{point clouds}
\cite{lhuillier2005quasidense, chen2019pointbased}, \textit{3D meshes}
\cite{fua1995object}, and \textit{depth maps}
\cite{Campbell2008UsingMH, Galliani2015fusibile,
  Johannes2016pixelwise, xu2019multiscale, gu2019casmvsnet,
  yao2018mvsnet, YU2021AAcostvolume, ding2022transmvsnet,
  peng2022rethinkingMVS, Cheng2019USCNet}. Depth map-based methods can
 further be categorized into either traditional techniques based on
feature detection and solving for geometric constraints
\cite{Campbell2008UsingMH, Galliani2015fusibile,
  Furukawa2010AccurateDA, Johannes2016pixelwise}, or learning-based
methods \cite{xu2019multiscale, gu2019casmvsnet, yao2018mvsnet,
  YU2021AAcostvolume, ding2022transmvsnet, peng2022rethinkingMVS,
  Cheng2019USCNet}. The latter have become very popular in
the last few years.

Among the learning-based techniques, MVSNet \cite{yao2018mvsnet}
formulates a single-stage MVS pipeline by encoding camera parameters
via differential homography to build 3D cost volumes. It requires a huge amount of
memory and computation as it uses 3D U-Nets \cite{olaf2015unet}
to regularize the cost volume. Subsequent work has taken two main
approaches to alleviate this problem: RNNs
\cite{yao2019recurrent, wei2021aa, yan2020dynamicfusion, xu2021nonlocalrecurrent} and 
coarse-to-fine multi-stage methods \cite{xu2019multiscale,
  gu2019casmvsnet, YU2021AAcostvolume, ding2022transmvsnet,
  peng2022rethinkingMVS, Cheng2019USCNet}. 

Among the RNN-based methods, R-MVSNet \cite{yao2019recurrent}
sequentially regularizes the 2D cost maps along the depth direction
via gated recurrent units. AA-RMVSNet \cite{wei2021aa} slices the cost
volume along $D$ depth hypotheses and regularizes the horizontal and
vertical components using CNN and ConvLSTMCells, respectively. Xu
et al. \cite{xu2021nonlocalrecurrent} use RNNs to model global
dependencies with non-local depth interactions. Yan et
al. \cite{yan2020dynamicfusion} couple LSTM and U-Net architectures to
regularize multi-scale information.

Coarse-to-fine multi-stage methods \cite{ding2022transmvsnet,
  peng2022rethinkingMVS, Luo2019Pmvsnet, yao2019recurrent,
  chen2019pointbased, gu2019casmvsnet, yang2019CVPMVS,
  Cheng2019USCNet, wei2021aa} have  significantly improved the
quality of depth estimates and point cloud reconstructions. They  initially predict a low-resolution
(coarse) depth map and then progressively refine it.
For example,
inspired by
other coarse-to-fine methods \cite{tonioni2018realtime,
  wang2018anytimestereo, yin2018hierarchical}, CasMVSNet
\cite{gu2019casmvsnet} presents a multi-stage formulation of
single-stage MVSNet \cite{yao2018mvsnet},
TransMVSNet \cite{ding2022transmvsnet} focuses on feature matching to
improve performance over CasMVSNet, UniMVSNet \cite{peng2022rethinkingMVS} uses 
unified loss formulation to further improve over CasMVSNet. 
CVP-MVSNet \cite{yang2019CVPMVS} builds a cost volume pyramid in a
coarse-to-fine manner.
UCS-Net \cite{gu2019casmvsnet} uses an adaptive thin volume
module that uses a smaller number of hypothesis planes to efficiently
partition the local depth range within learned small
intervals. TransMVSnet \cite{ding2022transmvsnet} uses transformer
based \cite{vaswani2017attention, angelos2020linearattention} feature
matching to promote similarity in the extracted features. UniMVSNet
\cite{peng2022rethinkingMVS} unifies the advantages of regression and
classification methods by designing a unified focal loss in a
multi-stage framework.

While all these methods improve the performance
of the multi-stage MVS pipeline by improving specific portions, none
of them explicitly models multi-view geometric cues during the learning
process. Consequently, during training these models depend  on the limited
geometric cues available from multiple source views and the cost function
formulation.
Xu and Tao \cite{xu2019multiscale} present a multi-scale geometric
consistency-guided MVS method that uses multi-hypothesis joint view
selection to leverage structured region information to sample better
candidate hypotheses. They hypothesize that the upsampled depth maps
of source images can geometrically constrain these estimates, and use reprojection error \cite{Johannes2016pixelwise,
  zhang2008consistentvideo} to indicate this consistency. In this
paper, we use forward-backward reprojection with multiple source
views to check the geometric consistency of depth estimates and to generate
per-pixel penalties for geometrically inconsistent pixels.

\begin{figure*}[t]
\vspace{-5pt}
\begin{center}
    \includegraphics[width=1\textwidth, height=14\baselineskip]{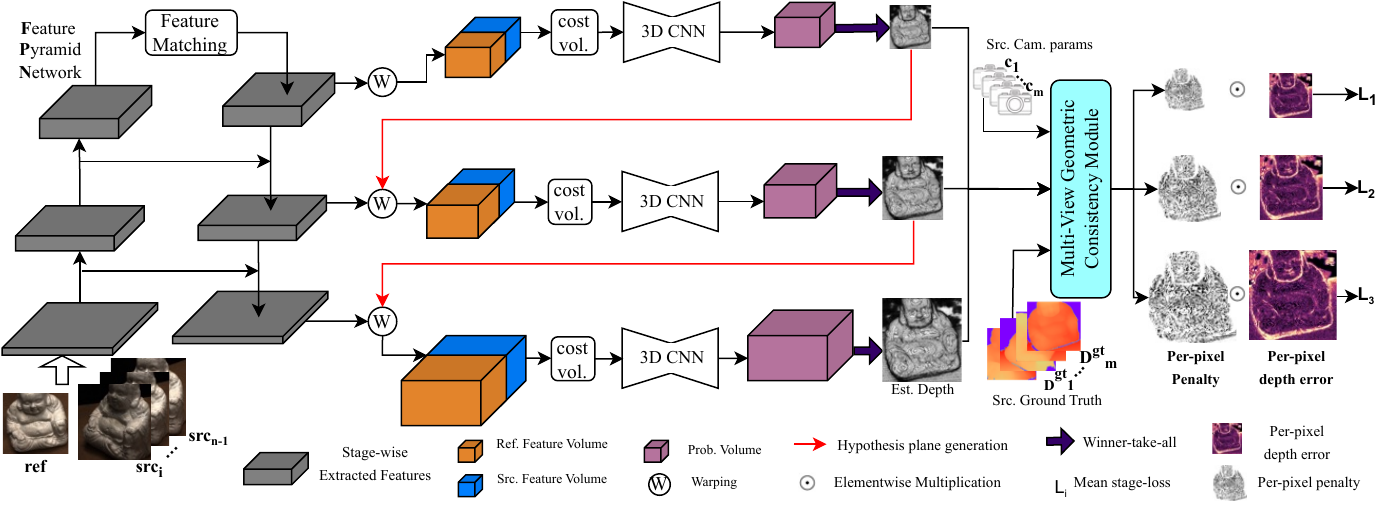}
    \vspace{-14pt}
    \caption{The GC-MVSNet architecture. The GC module is applied at the end of each stage. It takes the estimated reference view depth, $M$ source view ground truths and their camera parameters to perform a multi-view geometric consistency check. It generates a per-pixel penalty ($\xi_p$) for reference view, which is element-wise multiplied with per-pixel depth error ($\xi_d$) to generate stage loss $L_i$. $\xi_d$ is calculated using cross-entropy loss. All stage losses are added to produce the final loss.}
    \label{fig:gc-mvsnet architecture}
    \vspace{-25pt}
\end{center}
\end{figure*}

\section{Methodology}\label{sec:methodology}

Our goal is to take
$N$ views as input, including a
reference image \textbf{$I_0 \in \mathbb{R}^{H\times W \times 3}$} and
its paired $N$-1 source view images \{\textbf{$I_i$}\}$^{N-1}_{i=1}$,
along with the corresponding camera parameters $c_0, ..., c_N$, 
and then   to estimate the reference view depth
map ($D_0$) as the output.

\subsection{Network Overview}\label{sec:network-overview}

\cref{fig:gc-mvsnet architecture} shows the architecture of our approach, which we call
Geometric Consistency MVSNet (or GC-MVSNet). We use a deformable convolution-based
\cite{Dai2017deformableCNN}  feature pyramid network (FPN) \cite{Lin2016featurepyramid}
architecture (Sec. \ref{sec:other-modification}) to extract features
from input images in a coarse-to-fine manner in three stages.
 At each stage, we build a correlation-based cost volume
of shape $H' \times W'\times D'_h \times 1$ using feature
maps of shape $N \times H' \times W' \times C$,
where $H'$, $W'$, and
$C$ denote the height, width, and number of channels of a given stage, and
$D'_h$ is the number of depth
hypotheses at the corresponding stage.
The cost volume is
regularized with a cost regularization network.
We use a
winner-takes-all strategy to estimate the depth
map $D_{0}$ at each stage. At only the coarse stage, we apply feature matching
\cite{ding2022transmvsnet} with linear attention
\cite{angelos2020linearattention, ding2022transmvsnet} to leverage
global context information within and between reference and source
view features. 

We employ the GC module at each stage. The GC module
checks the geometric consistency of each pixel in $D_{0}$ across $M$
source views and generates $\xi_p$
(Sec. \ref{sec:multi-source-GC-module}), a pixel-wise 
factor that is
multiplied with the per-pixel depth error ($\xi_d$), calculated using a
cross-entropy function. It penalizes each pixel in $D_0$ for its
inconsistency across $M$ source views to accelerate geometric cues
learning during training. TransMVSNet \cite{ding2022transmvsnet} trained with cross-entropy loss (TransMVSNet-B) is our baseline; see Table \ref{table:performance-comparison-with-GC-module} for different stages of GC-MVSNet.

\vspace{-3pt}
\subsection{Multi-View Geometric Consistency Module}\label{sec:multi-source-GC-module}

GC-MVSNet estimates reference depth maps at three stages with
different resolutions. At each stage, the GC module takes $D_0$, $M$
source view ground truths $D^{gt}_1, ... D^{gt}_{M}$, and their camera
parameters $c_0, ... c_M$ as input (see Alg. \ref{alg:gc-algorithm}). The GC module is then initialized
with a \textit{geometric inconsistency mask sum} (or $mask\_sum$) of
zero at each stage. This mask sum accumulates the inconsistency of
each pixel across the $M$ source views. For each source view, the GC
module performs \textit{forward-backward reprojection}  of $D_0$
to generate the penalty and then adds it to the mask sum.

Forward-backward reprojection (FBR), as shown in Alg. \ref{alg:forward-backward-reprojection}, is a crucial three-step process. First,
we project each pixel $P_{0}$ of $D_{0}$ to
its $i^{th}$ neighboring source view using intrinsic ($K_R, K_S$) and extrinsic ($E_R, E_S$) camera 
parameters to obtain corresponding pixel $P'_{i}$, and denote the corresponding depth map as $D_{(R \rightarrow S)}$. Second, we 
similarly remap $D_i^{gt}$ 
to obtain $D_{S_{remap}}$.
Finally, 
we back project $D_{S_{remap}}$ to the reference view using intrinsic and extrinsic camera parameters 
to obtain $D''_{P''_{0}}$ (see Alg. \ref{alg:forward-backward-reprojection}). $D_0$ and $D''_{P''_{0}}$ represent
the depth values of pixels $P_0$ and
$P''_{0}$ \cite{Hartley2012Multi-view-geometry}. With $P''_{0}$
and $D''_{P''_{0}}$, we calculate the pixel displacement error (PDE)
and relative depth difference (RDD). PDE is the $L_2$ norm between
$P_0$ and $P''_{0}$ and RDD is the absolute value difference between
$D''_{P''_{0}}$ and $D_0$ relative to $D_0$ as shown in Alg. \ref{alg:gc-algorithm}.

\begin{algorithm}[t]
\footnotesize
\begin{algorithmic}
    \State \textbf{Inputs:} $D_0, c_0, D^{gt}_{i}, c^{gt}_{i}, D_{pixel}, D_{depth}$
    \State \textbf{Output:} $per\_pixel\_penalty$
    \State \textbf{Require} $M \ge N$
    \State $mask\_sum \gets 0$
    \State $D \gets {D_1^{gt}, ...D_M^{gt}}$
    \State $c \gets {c_1^{gt}, ...c_M^{gt}}$
    \For{$D_i^{gt},c_i^{gt}$ in $zip(D,c)$} 
    \State $D''_{P''_{0}}, P''_{0} \gets FBR(D_0,c_0,D_i^{gt},c_i^{gt})$ \Comment{Alg. \ref{alg:forward-backward-reprojection}}
    \State $PDE \gets {\vert\vert P_0 - P''_{0} \vert\vert_2 }$
    \State $RDD \gets {\nicefrac{1}{D_0}\vert\vert D''_{P''_{0}} - D_0 \vert\vert_1}$
    \State $PDE_{mask} \gets {PDE > D_{pixel}}$
    \State $RDD_{mask} \gets {RDD > D_{depth}}$
    \State $mask \gets PDE_{mask} \lor RDD_{mask}$
    \If{$mask > 0$}
        \State $mask \gets 1$ 
    \Else
        \State $mask \gets 0$
    \EndIf
    \State $mask\_sum \gets mask\_sum + mask $
    \EndFor \\
$per\_pixel\_penalty \gets 1 + mask\_sum/M$
\end{algorithmic}
\caption{Geometric Consistency Check Algorithm}
\label{alg:gc-algorithm}
\end{algorithm}

For each stage, we generate two binary masks of inconsistent pixels, $PDE_{mask}$ and $RDD_{mask}$,
by applying thresholds
$D_{pixel}$ and $D_{depth}$, and then take a logical-OR of the two to preduce a single mask
of inconsistent pixels.
These inconsistent pixels are assigned a value $1$ and all
other pixels, including the consistent and the out-of-scope pixels,
are assigned $0$ to form a penalty mask. This penalty mask
is then added to the mask sum (Alg. \ref{alg:gc-algorithm}),
which accumulates the penalty mask for each of the $M$ source views to generate a final
mask sum with values $\in [0, M]$.
Each pixel value indicates the number of inconsistencies of the pixel
across the $M$ source views.

\begin{figure}[t]
\begin{center}
    \vspace{-7pt}
   \includegraphics[width=0.9\linewidth]{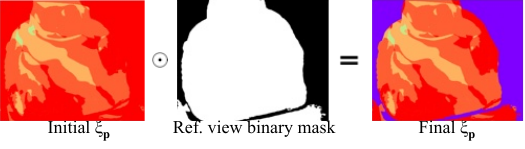}
    \vspace{-10pt}
    \caption{The final $\xi_p$ is the outcome of elementwise multiplication ($\odot$) of initial $\xi_p$ and reference view mask. It restricts the penalties within the reference view mask.}
    \label{fig:geo-weight-masking}
\end{center}
\vspace{-25pt}
\end{figure}

From this mask sum, we then generate the inconsistency penalty $\xi_p$
for each pixel. Our initial approach generated $\xi_p$ by dividing the mask sum by 
$M$ to normalize within the $[0,1]$.
However, we found that using $\xi_p$ itself for element-wise
multiplication reduces the contribution of perfectly consistent (zero
inconsistency) pixels to zero, preventing further improvement of such
pixels.  To avoid this, we add $1$ so that elements of $\xi_p$ are in
$[1,2]$. A reference view binary mask is applied on initial $\xi_p$ to
generate the final $\xi_p$, as shown in Fig. \ref{fig:geo-weight-masking}.

\begin{algorithm}[t]
\footnotesize
\begin{algorithmic}
    \State \textbf{Inputs:} $D_0,c_0,D_i^{gt},c_i^{gt}$
    \State \textbf{Output:} $D''_{P''_{0}}, P''_{0}$
    \State $K_R, E_R \gets c_0$;  $K_S, E_S \gets c_i^{gt}$
    \State $D_{(R \rightarrow S)} \gets K_S \cdot E_S \cdot E_R^{-1} \cdot K_R^{-1} \cdot D_0$ \Comment{Project}
    \State $X_{D_{(R \rightarrow S)}}, Y_{D_{(R \rightarrow S)}} \gets D_{(R \rightarrow S)}$ 
    \State $D_{S_{remap}}\gets REMAP(D_i^{gt}, X_{D_{(R \rightarrow S)}}, Y_{D_{(R \rightarrow S)}})$  \Comment{Remap}
    \State $D''_{P''_{0}} \gets K_R \cdot E_R \cdot E_S^{-1} \cdot K_S^{-1} \cdot D_{S_{remap}}$         \Comment{Back project}
    \State $P''_{0} \gets (X_{D''_{P''_{0}}}, Y_{D''_{P''_{0}}})$
\end{algorithmic}
\caption{Forward Backward Reprojection (FBR)}
\label{alg:forward-backward-reprojection}
\end{algorithm}

\vspace{-10pt}
\paragraph{Occlusion and its impact.}
Occluded pixels naturally arise in multi-view stereo, since 3D points
are often not visible in all views. These occluded pixels have a
major impact on geometric constraints, since the 
reference view pixels whose corresponding 3D points
are occluded are penalized as inconsistent.
It is thus important to prevent occluded pixels from dominating the
geometric consistency losses.  While occlusion is sometimes modeled
explicitly \cite{kang2001occlusion, nakamura1996stereoocclusion}, we
found that our approach is naturally robust to occlusion because of 
the following three considerations. First, we select the closest $M$ source views as defined in
MVSNet \cite{yao2018mvsnet} to minimize the number of occluded pixels
in different source views. Then, during FBR, we remap $D_i^{gt}$ to 
obtain $D_{S_{remap}}$ and back project it as shown in Alg. \ref{alg:forward-backward-reprojection}. Remapping and back
projection largely handles extreme cases of occlusion (see Appendix A
in Supplemental Material). Finally, we apply
reference view binary mask on $\xi_p$,
Fig. \ref{fig:geo-weight-masking}, to restrict penalties only to valid
reference view pixels. The combination of these steps helps us deal
with occluded pixels and loss explosion.

\vspace{-5pt}
\subsection{Cost Function}\label{sec:cost-function}

Most learning-based MVS methods \cite{gu2019casmvsnet, yang2019CVPMVS,Zhang2020visibility} 
treat depth estimation as a regression problem and use an $L_1$ loss between prediction and ground truth. Following AA-RMVSNet \cite{wei2021aa} and UniMVSNet \cite{peng2022rethinkingMVS}, we treat depth 
estimation as a classification problem and adopt a cross-entropy loss formulation 
from AA-RMVSNet \cite{wei2021aa} (see \cite{peng2022rethinkingMVS} for 
relative advantages of regression and classification approaches.) The pixelwise 
depth error $\xi_d$ is calculated at each stage,

\vspace{-8pt}
\begin{equation}
    \xi_d = \mathcal{D}(D^{gt}_{0}, D_0)
\end{equation}
\vspace{-5pt}

\noindent where $D^{gt}_{0}$ is the reference ground truth and $D_0$
is the reference depth estimate. $\mathcal{D}$ denotes the cross-entropy function 
modified to produce per-pixel depth error between $D^{gt}_{0}$ and $D_0$.
We further
enhance the one-hot supervision by penalizing each pixel for its
inconsistency across different source views. This is implemented using
element-wise multiplication ($\odot$) between $\xi_d$ and $\xi_p$ at
each stage. The mean stage loss, $L_i$, is calculated as,

\vspace{-14pt}
\begin{equation}
\begin{split}
    {L_i}_{(stage)} & = mean (\xi_p \odot \xi_d)\\
    \mathcal{L}_{total} & = \alpha.L_1 + \beta.L_2 + \gamma.L_3
\end{split}
\end{equation}
\vspace{-10pt}

\noindent where ${L_i}_{(stage)}$ is the mean stage loss and $\mathcal{L}_{total}$ is the total loss. 
$\alpha, \beta$ and $\gamma$ are the stage-wise weights.
This formulation of cost function
with pixel-level inconsistency penalty explicitly forces the model to learn to produce multi-view geometrically-consistent depth maps.

\subsection{Other Modifications}\label{sec:other-modification}

Besides the geometric consistency module, we made two other major modifications to the MVS pipeline.
First, while keeping the feature
extraction network architecture as FPN, we
replaced its regular convolutional layers with deformable layers
\cite{Dai2017deformableCNN, zhu2018deformableV2}. Deformable layers
are known to adjust their sampling locations based on model
requirements \cite{Dai2017deformableCNN, zhu2018deformableV2}. This
helps extract better features for accelerated learning.

Second, most MVS methods \cite{gu2019casmvsnet, ding2022transmvsnet,
  peng2022rethinkingMVS, wei2021aa, Zhang2020visibility,
  weilharter2021highresMVS} use batch normalization 
\cite{ioffe2015batchnorm} and batch synchronisation during training.
As observed in \cite{ioffe2015batchnorm}, batch normalization provides more consistent
and stable training with large batch sizes, but it is inconsistent and
has a degrading effect on training with smaller batches. MVS
methods are restricted to very small batch sizes, often $1$, due to
large memory requirements. Thus, we replaced batch normalization with 
group normalization  layers
\cite{he2018groupnorm} of group size $4$ across the network. Group normalization performs normalization across a 
number of channels that is independent of the
number of examples in a batch \cite{he2018groupnorm}. We also implement weight standardization
\cite{qiao2019weightstandardization} for all layers in the
network. With these modifications, we achieve stable and reproducible
training (see Appendix D in Supplemental Material).

\vspace{-5pt}
\section{Experiments}\label{sec:experiments}

We evaluate on three datasets with different
complexities. \textbf{DTU} \cite{jensen2014dtu} is an indoor dataset
that contains $128$ scenes with $49$ or $64$ views under $7$ lighting
conditions and predefined camera trajectories. We follow MVSNet
\cite{yao2018mvsnet} for training, validation, and test
splits. \textbf{BlendedMVS}\cite{yao2019blended} is a
large-scale synthetic dataset with $113$ indoor and outdoor scenes. It
has 106 training scenes and $7$ validation scenes. \textbf{Tanks and
  Temples} \cite{Knapitsch2017tnt} is collected from a more
complicated and realistic scene, and contains $8$ intermediate and $6$
advanced scenes. DTU and TnT evaluate using point clouds while BLD
evaluates on depth maps.


\begin{table}[t]
  \begin{center}
    {\footnotesize{
\begin{tabular}{clccc}
\toprule
& Method  & Acc $\downarrow$ & Comp $\downarrow$ & Overall $\downarrow$ \\
\cmidrule{2-5}
\parbox[t]{2mm}{\multirow{4}{*}{\rotatebox[origin=c]{90}{Traditional}}} & Furu \cite{Furukawa2010AccurateDA} & 0.613 & 0.941 & 0.777 \\
& Tola \cite{tola2011largescale} & 0.342 & 1.190 & 0.766 \\
& Gipuma \cite{Galliani2015fusibile}  & \textbf{0.283} & 0.873 & 0.578 \\
& COLMAP \cite{Johannes2016pixelwise}  &  0.400 & 0.664 & 0.532 \\ 
\cmidrule{2-5}
\parbox[t]{2mm}{\multirow{16}{*}{\rotatebox[origin=c]{90}{Learning-based}}}& SurfaceNet \cite{Ji2017surfacenet} & 0.450 & 1.040 & 0.745 \\
& MVSNet \cite{yao2018mvsnet}  & 0.396 & 0.527 & 0.462 \\
& P-MVSNet \cite{Luo2019Pmvsnet} &  0.406 & 0.434 & 0.420 \\
& R-MVSNet \cite{yao2019recurrent}  &  0.383 & 0.452 & 0.417 \\
& Point-MVSNet \cite{chen2019pointbased}  & 0.342 & 0.411 & 0.376 \\
& CasMVSNet \cite{gu2019casmvsnet}   & 0.325 & 0.385 & 0.355 \\
& CVP-MVSNet \cite{yang2019CVPMVS}  & \underline{0.296} & 0.406 & 0.351 \\
& UCS-Net \cite{Cheng2019USCNet}  & 0.338 & 0.349 & 0.344 \\
& AA-RMVSNet \cite{wei2021aa}  & 0.376 & 0.339 & 0.357 \\
& UniMVSNet \cite{peng2022rethinkingMVS}  & 0.352 & 0.278 & 0.315 \\
& TransMVSNet \cite{ding2022transmvsnet}  & 0.321 & 0.289 & 0.305\\
& GBi-Net$^{*}$ \cite{mi2022gbi} & 0.312 & 0.293 & \underline{0.303} \\
& MVSTER \cite{wang2022mvster} & 0.350 & \underline{0.276} & 0.313 \\
& \textbf{GC-MVSNet} (ours) & 0.330 & \textbf{0.260} & \textbf{0.295}\\
\cmidrule{2-5}
& \textcolor{Gray}{GBi-Net \cite{mi2022gbi}} & \textcolor{Gray}{0.315} & \textcolor{Gray}{0.262} & \textcolor{Gray}{0.289} \\
& \textcolor{Gray}{\textbf{GC-MVSNet} (ours)} & \textcolor{Gray}{0.323} & \textcolor{Gray}{\textbf{0.255}} & \textcolor{Gray}{0.289}\\
\bottomrule
\end{tabular}
}}
\vspace{-6pt}
\caption{Quantitative results on DTU evaluation set at 864 $\times$ 1152 resolution. Accuracy (Acc), completeness (comp) and overall are in $mm$. $*$ means that GBiNet is re-tested with the same post-processing threshold to all scans for fair comparison with other methods. Gray font shows the methods that use scan-specific thresholds for evaluation. \textbf{Bold} and \underline{underline} represents first and second place, respectively.}
\label{Table:DTU-model-comparison}
\vspace{-22pt}
\end{center}
\end{table}

\begin{table}[t]
  \begin{center}
    {\footnotesize{
\begin{tabular}{lccc}
    \toprule
    Method  & EPE $\downarrow$ & $e_1 \downarrow$ & $e_3 \downarrow$\\
    \midrule
    MVSNet \cite{yao2018mvsnet} & 1.49  &  21.98 &  8.32 \\
    CasMVSNet \cite{gu2019casmvsnet} &  1.43 & 19.01  &  9.77 \\
    CVP-MVSNet \cite{yang2019CVPMVS} &  1.90 & 19.73  & 10.24  \\
    Vis-MVSNet \cite{Zhang2020visibility} & 1.47  & 15.14  &  5.13 \\
    EPP-MVSNet \cite{ma2021eppMVS} & 1.17  & 12.66 &  6.20 \\
    TransMVSNet \cite{ding2022transmvsnet} & \underline{0.73} & \underline{8.32} & \underline{3.62} \\
    \midrule
    \textbf{GC-MVSNet} (ours) & \textbf{0.48}  & \textbf{7.48}  &  \textbf{2.78} \\
    \bottomrule
\end{tabular}
}}
\vspace{-6pt}
\caption{Quantitative comparison on BlendedMVS evaluation set. We follow evaluation steps described in \cite{darmon2021wildMVS}. \textbf{Bold} and \underline{underline} represents first and second place, respectively.}
\label{table:quantitative-comparison-on-blended}
\vspace{-25pt}
\end{center}
\end{table}

\vspace{-3pt}
\subsection{Implementation Details}\label{sec:implementation-details}

Following the general practice \cite{ding2022transmvsnet}, we first train
and evaluate our model on DTU. Then, we finetune on BlendedMVS to
evaluate on Tanks and Temples. For training on DTU, we set the number
of input images $N=5$ and image resolution as $512 \times 640$. The
depth hypotheses are sampled from $425mm$ to $935mm$ for
coarse-to-fine regularization with the number of plane sweeping depth
hypotheses for the three stages set to 48, 32, and 8. The corresponding depth
interval ratio (DIR) is set as 2.0, 0.8, and 0.4. The model is trained
with Adam \cite{Kingma2014AdamAM} for 9 epochs with an initial
learning rate ($LR_{DTU}$) of 0.001, which decays by a factor of 0.5
once after $8^{th}$ epoch. For the Geometric Consistency (GC) module, we use $M$=8 and set
the stage-wise thresholds $D_{pixel}$ as 1, 0.5, 0.25 and ${D_{depth}}$ as
0.01, 0.005, 0.0025. We use $\alpha$=$\beta$=1 and $\gamma=2$ for all experiments. We train our model with a batch size of 3 on 8
NVIDIA RTX A6000 GPUs for about 9 hours.

\begin{figure*}[t]
\begin{center}
    \vspace{-10pt}
    \includegraphics[width=0.95\textwidth,height=16\baselineskip]{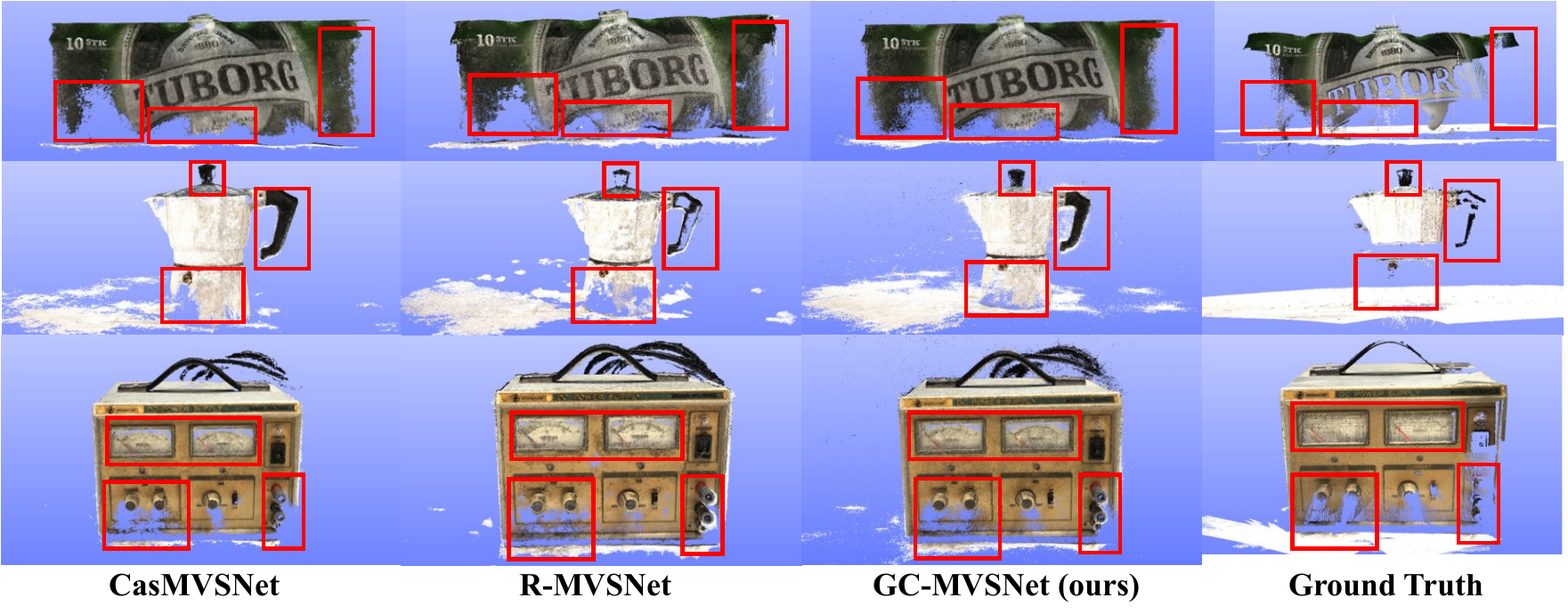}
    \vspace{-15pt}
    \caption{Visual comparison of reconstructed point clouds of GC-MVSNet with CasMVSNet \cite{gu2019casmvsnet}, R-MVSNet \cite{yao2019recurrent} and Ground truths. Our method obtains a more complete point cloud. See Appendix H in Supplemental Material for all point clouds.}
    \label{fig:point-cloud-comparison}
    \vspace{-17pt}
\end{center}
\end{figure*}

\begingroup
\setlength{\tabcolsep}{3pt} 
\begin{table*}[ht]
\footnotesize
  \begin{center}
    \begin{tabular}{lccccccccc|ccccccc}
    \toprule
    & \multicolumn{9}{c}{\textbf{Intermediate set}} & \multicolumn{7}{c}{\textbf{Advanced set}} \\
    \cmidrule{2-17}
    Method & \textbf{Mean $\uparrow$} & Fam. & Fra. & Hor. & Lig. & M60 & Pan. & Pla. & Tra. & \textbf{Mean $\uparrow$} & Aud. & Bal. & Cour. & Mus. & Pal. & Tem.\\
    \midrule
    COLMAP \cite{Johannes2016pixelwise} & 42.14 & 50.41 & 22.25 & 26.63 & 56.53 & 44.83 & 46.97 & 48.53 & 42.04 & 27.24 & 16.02 & 25.23 & 34.70 & 41.51 & 18.05 & 27.94\\
    P-MVSNet \cite{Luo2019Pmvsnet}& 55.62 & 70.04 & 44.64 & 40.22 & \textbf{65.20} & 55.08 & 55.17 & 60.37 & 54.29  & - & - & - & - & - & - & - \\
    R-MVSNet \cite{yao2019recurrent} & 50.55 & 73.01 & 54.56 & 43.42 & 43.88 & 46.80 & 46.69 & 50.87 & 45.25  & 29.55 & 19.49 & 31.45 & 29.99 & 42.31 & 22.94 & 31.10\\
    Point-MVSNet \cite{chen2019pointbased}& 48.27 & 61.79 & 41.15 & 34.20 & 50.79 & 51.97 & 50.85 & 52.38 & 43.06 & - & - & - & - & - & - & -\\
    CasMVSNet \cite{gu2019casmvsnet} & 56.84 & 76.37 & 58.45 & 46.26 & 55.81 & 56.11 & 54.06 & 58.18 &  49.51 & 31.12 & 19.81 & 38.46 & 29.10 & 43.87 & 27.36 & 28.11\\
    CVP-MVSNet \cite{yang2019CVPMVS}& 54.03 & 76.50 & 47.74 & 36.34 & 55.12 & 57.28 & 54.28 & 57.43 & 47.54  & - & - & - & - & - & - & - \\
    UCS-Net \cite{Cheng2019USCNet}& 54.83 & 76.09 & 53.16 & 43.03 & 54.00 & 55.60 & 51.49 & 57.38 & 47.89  & - & - & - & - & - & - & - \\
    AA-RMVSNet \cite{wei2021aa} & 61.51 & 77.77 & 59.53 & 51.53 & \underline{64.02} & \underline{64.05}& 59.47 & \underline{60.85} & 54.90  & 33.53 & 20.96 & 40.15 & 32.05 & 46.01 & 29.28 & 32.71\\
    UniMVSNet \cite{peng2022rethinkingMVS} & \textbf{64.36} & \textbf{81.20} & \underline{\underline{66.43}} & \underline{\underline{53.11}} & \underline{\underline{63.46}} & \textbf{66.09} & \textbf{64.84} & \textbf{62.23} & \underline{\underline{57.53}}  & \textbf{38.96} & \underline{28.33} & \underline{\underline{44.36}} & \textbf{39.74} & \textbf{52.89} & \underline{\underline{33.80}} & 34.63 \\
    TransMVSNet \cite{ding2022transmvsnet} & \underline{63.52} & \underline{80.92} & 65.83 & \textbf{56.94} & 62.54 & \underline{\underline{63.06}} & \underline{60.00} & 60.20 & \textbf{58.67}  & 37.00 & 24.84 & \underline{44.59} & 34.77 & 46.49 & \underline{34.69} & \underline{\underline{36.62 }}\\
    GBi-Net \cite{mi2022gbi}  & 61.42 & 79.77 & \textbf{67.69} & 51.81 & 61.25 & 60.37 & 55.87 & \underline{\underline{60.67}} & 53.89 & 37.32 & \textbf{29.77} & 42.12 & \underline{\underline{36.30}} & 47.69 & 31.11 & 36.39  \\
    MVSTER \cite{wang2022mvster}  & 60.92 & 80.21 & 63.51 & 52.30 & 61.38 & 61.47 & 58.16 & 58.98 & 51.38 & \underline{\underline{37.53}} & \underline{\underline{26.68}} & 42.14 & 35.65 & \underline{\underline{49.37}} & 32.16 & \textbf{39.19}  \\
    \midrule
    GC-MVSNet(ours) & \underline{\underline{62.74}}  & \underline{\underline{80.87}}  & \underline{67.13}  &  \underline{53.82} & 61.05  & 62.60  & \underline{\underline{59.64}}  &  58.68 &  \underline{58.48} & \underline{38.74} & 25.37  &  \textbf{46.50} & \underline{36.65} & \underline{49.97} & \textbf{35.81} & \underline{38.11}\\ 
    \bottomrule
    \end{tabular}
    \vspace{-5pt}
    \caption{Quantitative results on intermediate and advanced sets of Tanks and Temples \cite{Knapitsch2017tnt}. \textbf{Bold}, \underline{single-underline}, \underline{\underline{double-underline}} represent first, second and third places, respectively.}
    \label{table:quantitative-comparison-tanks-and-temples}
    \vspace{-25pt}
\end{center}
\end{table*}
\endgroup

\subsection{Experimental Performance}\label{sec:experimental-performance}

\vspace{-3pt}
\paragraph{Evaluation on DTU.}
On DTU, we generate depth maps with $N$=5 at an input
resolution of $864\times1152$. We slightly adjust the depth interval ratio (DIR) to $1.6,
0.7, 0.3$ to accommodate the resolution change (more on DIR in
Appendix C in Supplemental Material) and use the Fusibile algorithm \cite{Galliani2015fusibile} for
depth fusion.
Table \ref{Table:DTU-model-comparison} shows quantitative
evaluations, where accuracy is the mean
absolute distance in $mm$ from the reconstructed point cloud to the
ground truth point cloud, completeness measures the opposite
(see Appendix F in Supplemental Material), and overall is the average of
these metrics, indicating the overall performance of the
models. We find that GC-MVSNet achieves the best overall score as well as the best
completeness score, when compared to nearly two dozen previous and state-of-the-art techniques.
A qualitative evaluation is presented in Fig. \ref{fig:point-cloud-comparison} on a few
sample MVS problems.
We find that our model
generates denser and more complete point clouds.

\vspace{-10pt}
\paragraph{Evaluation on BlendedMVS.}
Unlike DTU and Tanks and Temples, evaluation on Blended MVS is usually measured
as the quality of depth maps, not the quality of point clouds.
We set $N$=5, $M$=8, image
resolution as $576 \times 768$, and number of depth planes $D$=128, and
finetune for 10 epochs with one-tenth the learning rate we used for DTU ($\frac{1}{10} LR_{DTU}$).
We follow \cite{darmon2021wildMVS} for evaluation process.

Table \ref{table:quantitative-comparison-on-blended} presents the results of 
our quantitative evaluation, using three metrics:
Endpoint error (EPE) is the average $L_1$
distance between the estimated and the ground truth depth values, and
$e_1$ and $e_2$ are the ratio of number of pixels with $L_1$ error
larger than 1\textit{mm} and 3\textit{mm}, respectively.
The significant improvement in depth map estimates corroborates that providing explicit geometric cues during training
helps the model learn about multi-view geometric consistency while requiring much less training iteration. See Appendix H of the Supplemental Material for point clouds.

\begin{figure*}[t]
\begin{center}
    \vspace{-12pt}
    \includegraphics[width=0.95\textwidth,height=13\baselineskip]{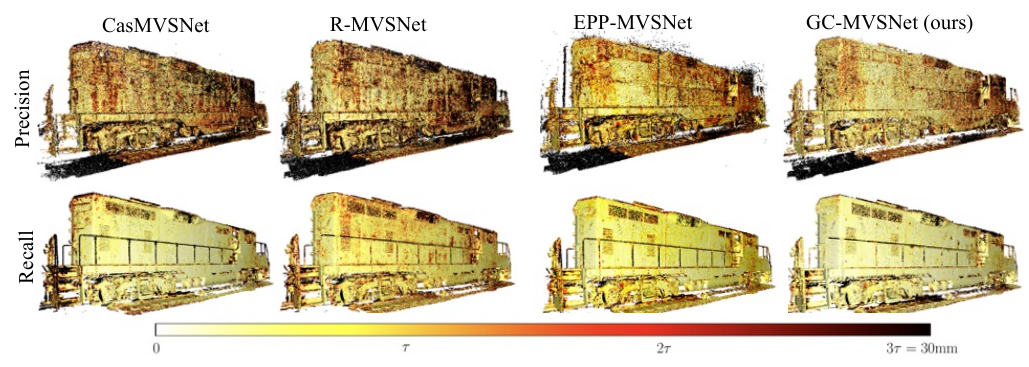}
    \vspace{-13pt}
    \caption{Precision and recall comparison with other methods \cite{gu2019casmvsnet, yao2019recurrent, ma2021epp} for Train on Tanks and Temples benchmark. $\tau$ is the scene-relevant distance threshold. Darker regions indicate larger error encountered with regard to $\tau$. GC-MVSNet shows visual improvements with brighter regions for precision as well as recall metric.}
    \label{fig:tnt-point-cloud-main-paper}
    \vspace{-25pt}
\end{center}
\end{figure*}

\vspace{-12.5pt}
\paragraph{Evaluation on Tanks and Temples.}
We also test the
performance of our model on an outdoor dataset with the Tanks and Temples benchmark. To
adapt to this change, we first finetune our model on BlendedMVS and then
evaluate on the intermediate and advanced test sets of Tanks and Temples. We use an image
resolution of $576 \times 768$, $N$=7, $M$=10, one-tenth the learning rate of DTU ($\frac{1}{10}LR_{DTU}$),
and $D$=192 for finetuning. We finetune the model for 12 epochs.
The camera parameters and neighboring view
selection are used as in R-MVSNet \cite{yao2019recurrent} 
and follow
evaluation steps described in CDS-MVSNet
\cite{giang2022curvatureguided}.

Table
\ref{table:quantitative-comparison-tanks-and-temples} presents our
quantitative comparison of different methods. GC-MVSNet achieves the 
third highest spot on the intermediate set and the second highest spot on advanced set
evaluation. Fig. \ref{fig:tnt-point-cloud-main-paper} shows point clouds visualizing precision
and recall comparisons with other MVS methods. See Appendix H in Supplemental Material for point clouds.

\subsection{Ablation Study}\label{sec:ablation-study}

Having demonstrated the efficacy of our proposed approach relative to
the state of the art, we now conduct ablation studies 
to evaluate the importance of the various components of our model.

\vspace{-10pt}
\paragraph{Range of $\xi_p$.}
$\xi_p$ is generated using the mask sum ($mask\_sum$ in Alg. \ref{alg:gc-algorithm}). 
It is the sum of penalties accumulated across the $M$ source views during multi-view geometric consistency check. At this stage, its elements take a discrete value between $0$ and $M$. Using mask sum as it is leads to very high penalty
per-pixel and consequently, very high loss value. Such a high loss value 
destabilizes the learning process. We control the magnitude of penalty by controlling the range of per-pixel penalty. 

We explore two different ranges to control the magnitude of $\xi_p$,
$[1,2]$ and $[1,3]$. To generate $\xi_p \in [1,2]$, we divide the mask
sum by $M$ and then add $1$. To
generate $\xi_p \in [1,3]$, we divide the mask by $\nicefrac{M}{2}$
and then add $1$. Table
\ref{table:range-of-per-pixel-penalty} shows the impact of these two
ranges of penalty for $M$=8. Since $\xi_p \in [1,2]$ produces best results,
we use it for all other experiments.

\vspace{-14pt}
\paragraph{Hyperparameters of GC module.}
The GC module has two types of hyperparameters, global and local. In this section,
we investigate the effect of these hyperparameters on our results.

The
global hyperparameter $M$ is the number of source views across which
the geometric consistency is checked, and is the same for all three
stages (coarse, intermediate, and refinement stages).  For training on
DTU, we vary the value of $M$ while keeping $N$=5, i.e. while the MVS
method uses only $4$ source views to estimate $D_0$, the GC module
checks the geometrical consistency of $D_0$ across $M$ source
views. It is important to note that the first $N$-1 out of $M$ source
views are exactly the same used by GC-MVSNet to estimate $D_0$. We
always keep $M \geq N-1$.

Table \ref{table:src views gc check ablation} presents a quantitative
comparison for different values of $M$ and the amount of training
iterations required for optimal performance of the model. At $M$=$N$-1= 4,
i.e. checking geometric consistency across the same number of source
views as used by GC-MVSNet to estimate $D_0$, the model performance
significantly improves with sharp decrease in training iteration requirements, 
as compared to our baseline TransMVSNet-B (Table
\ref{table:performance-comparison-with-GC-module}). As we increase the
value of $M$ from $4$ to $10$, the training iteration required by our model
further decreases. We find that at $M=8$, which is twice the number of
source views used by GC-MVSNet, it achieves its best performance.


The two local hyperparameters, $D_{pixel}$ and $D_{depth}$, are the
stage-wise thresholds applied to generate $PDE_{mask}$ and
$RDD_{mask}$ in Alg. \ref{alg:gc-algorithm}. These values are set to smaller values in the later (finer) stages,
providing a stricter
penalty to geometrically inconsistent pixels at finer
resolutions. Table \ref{table:gc-module-hyperparameter} shows the
overall performance of GC-MVSNet with a range of different $D_{pixel}$
and $D_{depth}$ thresholds. GC-MVSNet performance remains fairly consistent and it achieves its best performance
with $D_{pixel}$= 1, 0.5, 0.25 and $D_{depth}$= 0.01, 0.005, 0.0025. We
use these threshold values for all datasets throughout the paper.



\begin{table}[t]
  \begin{center}
    {\footnotesize{
\begin{tabular}{cccc}
\toprule
$\xi_p$ Range  & Acc$\downarrow$ & Comp$\downarrow$ & Overall$\downarrow$ \\
\midrule
$[1,3]$ & 0.331 & 0.270 & 0.3005 \\ 
$[1,2]$ & \textbf{0.330} & \textbf{0.260} & \textbf{0.295} \\
\bottomrule
\end{tabular}}}
\vspace{-6pt}
\caption{Impact of range of $\xi_p$ during training on DTU with $M$=8, $N$=5. Numbers are generated on DTU evaluation set.}
\label{table:range-of-per-pixel-penalty}
\vspace{-10pt}
\end{center}

  \begin{center}
    {\footnotesize{
\begin{tabular}{ccccc}
\toprule
M  & Acc$\downarrow$ & Comp$\downarrow$ & Overall$\downarrow$ & Opt. Epoch\\
\midrule
4 & 0.343 & 0.264  &  0.3035  & 12\\
5 & 0.342 & 0.271  & 0.3065   &  13\\
6 & \textbf{0.326} & 0.271  &  0.298  & 9 \\
7 & 0.332 &  0.270 &  0.301  & 10 \\
\textbf{8} & 0.330 &  \textbf{0.260} &  \textbf{0.295}  & 9 \\
9 & 0.328 &  0.280 & 0.304  & 9 \\
10 & 0.329 & 0.268  &  0.298 & 10 \\
\bottomrule
\end{tabular}}}
\vspace{-6pt}
\caption{Quantitative results on DTU evaluation set \cite{jensen2014dtu}. M is the number of source views used by the GC module for checking geometric consistency of reference view depth map. Training iteration requirement of the model decreases as M increases.}
\label{table:src views gc check ablation}
\end{center}
\vspace{-30pt}
\end{table}

\begin{table}[t]
  \begin{center}
    {\footnotesize{
\begin{tabular}{ccccccc}
    \toprule
    \multicolumn{3}{c}{$D_{depth}$} & \multicolumn{3}{c}{$D_{pixel}$} & \multirow{2}{*}{Overall$\downarrow$} \\
    \cmidrule{1-6}
    C & I & R & C & I & R &  \\
    \midrule
    0.04 & 0.03 & 0.02 & 4 & 3 & 2 & 0.302\\
    0.03 & 0.0225 & 0.015 & 3 & 2.25 & 1.5 & 0.302 \\
    0.02 & 0.015 & 0.01 & 2 & 1.5 & 1.0 & 0.298 \\
    0.01 & 0.008 & 0.006 & 1 & 0.8 & 0.6 & 0.303 \\
    \textbf{0.01} & \textbf{0.005} & \textbf{0.0025} & \textbf{1} & \textbf{0.5} & \textbf{0.25} & \textbf{0.295} \\
    0.008 & 0.003 & 0.002 & 0.8 & 0.3 & 0.2 & 0.303 \\
    0.005 & 0.002 & 0.001 & 0.5 & 0.2 & 0.1 & 0.3015 \\
    \bottomrule
\end{tabular}
}}
\vspace{-6pt}
\caption{Overall score on the evaluation set of DTU \cite{jensen2014dtu} for different values of $D_{depth}$ and $D_{pixel}$. $M$ is fixed at $8$. C, I, and R means Coarse, Intermediate and Refine stages.}
\label{table:gc-module-hyperparameter}
\vspace{-8pt}
\end{center}

    \vspace{-18pt}
  \begin{center}
    {\footnotesize{
\begin{tabular}{lllcccc}
\toprule
  & Methods & Loss & Other & GC & Overall$\downarrow$ & Epoch\\
\cmidrule{2-7}
\parbox[t]{1mm}{\multirow{6}{*}{\rotatebox[origin=c]{90}{GC as a plug-in}}} &\multirow{3}{*}{CasMVSNet} & $L_1$ & $\times$ & $\times$ & 0.355 & 16\\
 & & $L_1$ & $\checkmark$ & $\times$ & 0.357 & 16\\
 & & $L_1$ & $\times$ & $\checkmark$ & \textbf{0.335} & \textbf{11} \\
\cmidrule{2-7}
& \multirow{3}{*}{TransMVSNet} & FL & $\times$ & $\times$ & 0.305 & 16 \\
& & FL & $\checkmark$ & $\times$ & 0.322 & 16 \\
& & FL & $\times$ & $\checkmark$ & \textbf{0.303} & \textbf{8} \\
\midrule
\parbox[t]{2mm}{\multirow{3}{*}{\rotatebox[origin=c]{90}{Stages}}}& TransMVSNet-B & CE & $\times$ & $\times$ & 0.332 & 16 \\
& TransMVSNet-B & CE & $\checkmark$ & $\times$ & 0.328 & 16 \\
& TransMVSNet-B & CE & $\times$ & $\checkmark$ & 0.298 &  \textbf{8} \\
& GC-MVSNet & CE & $\checkmark$ & $\checkmark$ & \textbf{0.295} & 9 \\
\bottomrule
\end{tabular}
}}
\vspace{-6pt}
\caption{Performance comparison of different MVS methods with different modifications on DTU \cite{jensen2014dtu}. $L_1$, FL, CE and Others indicate $L_1$ loss, Focal loss \cite{Lin2017FocalLoss}, Cross-entropy loss \cite{wei2021aa} and other modifications from Sec. \ref{sec:other-modification}, respectively.}
\label{table:performance-comparison-with-GC-module}
\vspace{-30pt}
\end{center}
\end{table}

\vspace{-10pt}
\paragraph{GC module as a plug-in.}
Our Geometric Consistency module is generic can be integrated into
many different MVS pipelines.  To demonstrate this, we tested it with
two very different MVS pipeline, CasMVSNet \cite{gu2019casmvsnet} and
TransMVSNet \cite{ding2022transmvsnet}.  CasMVSNet treats depth
estimation as a regression problem, while TransMVSNet treats it as a
classification problem and uses winner-take-all to estimate the final
depth map. We purposefully choose different methods to show that the
GC module can perform well for both types of formulation. We 
compare the architectures of GC-MVSNet with TransMVSNet and CasMVSNet
in Sec. \ref{sec:model_comparison}.

Table \ref{table:performance-comparison-with-GC-module} presents the results,
showing the impact
of adding the GC module as well as the \textit{other} modifications (deformable convolution-based FPN with group-norm and weight-standardization)
in the
original pipeline. To observe the absolute impact of adding these
modifications, we do not change anything else in the original
pipelines. We observe in the table that applying only the \textit{other}
modification leads to degradation in  performance. It
indicates that the \textit{other} modification helps in
stabilizing the training process and promoting reproducibility, but
has no significant impact on the performance of the model on its
own. We also observe a sharp increase in model
performance and decrease in training iteration requirements after integrating
our GC module into the original pipeline. With GC, training the
CasMVSNet pipeline requires only $11$ epochs 
instead of $16$ epochs, while TransMVSNet (with GC module) requires only $8$ epochs
instead of $16$ epochs. This corroborates our
hypothesis that multi-view geometric consistency significantly reduces training computation
because it accelerates learning of geometric cues.

Table \ref{table:performance-comparison-with-GC-module} also shows
different stages of development of GC-MVSNet.  TransMVSNet-B uses TransMVSNet pipeline with cross-entropy loss, performs much worse than original TransMVSNet
\cite{ding2022transmvsnet} which uses focal loss. With only \textit{other} 
modifications, it slightly improves the overall performance of the model but does not
  impact the training iteration requirements. Only after applying the GC module,
  independently and with \textit{other} modifications, we see significant reduction
  in training iteration requirements as well as a significant improvement in the overall
  accuracy metric. This clearly shows the significance of multi-view multi-scale geometric consistency check in the GC-MVSNet pipeline.

\vspace{-8pt}
\section{Discussion}
\subsection{Comparison to Related Work}\label{sec:model_comparison}

\paragraph{GC-MVSNet vs.~TransMVSNet.}
TransMVSNet \cite{ding2022transmvsnet} uses regular 2D
convolution-based FPN (with batch-norm) for feature extraction and
employs adaptive receptive field (ARF) modules with deformable layers
after feature extraction. It trains using focal loss
\cite{Lin2017FocalLoss}. GC-MVSNet replaces the combination of regular
FPN and ARF modules with deformable FPN (with group-norm and
weight-standardization) for feature extraction. It trains with cross-entropy  
loss
and GC module for accelerated learning.

\vspace{-10pt}
\paragraph{GC-MVSNet vs.~CasMVSNet:}
CasMVSNet \cite{gu2019casmvsnet} proposes a coarse-to-fine regularization
technique. It uses regular 2D convolutions-based FPN for feature
extraction, generates variance-based cost volume and employ depth
regression to estimate $D_0$. The only similarity with our model is
that we also use coarse-to-fine regularization.

\vspace{-5pt}
\subsection{Limitations}
Like any other MVS method, GC-MVSNet require hyper-parameter tuning during learning. Hyperparameters like, depth interval ratio, number of stage-wise depth hypothesis, number of initial depth hypothesis, depth interval decay factor, etc. impacts model performance. The GC module hyperparameters, $D_{pixel}, D_{depth}$ and $M$, also require tuning to achieve its best performance. Along with the GC module hyperparameters, the quality of ground truth has also a direct impact on its performance as it uses source view ground truth depth maps for multi-view geometric consistency check.

\vspace{-8pt}
\section{Conclusion}

In this paper, we present a novel learning-based MVS pipeline,
GC-MVSNet, which explicitly models geometric consistency of reference
depth maps across multiple source views during training. To the best
of our knowledge, this is the first attempt to leverage multi-view
multi-scale geometric consistency check during the training
process. We show that the GC module is generic and can be plugged into
other MVS methods to accelerate their learning as well.  We perform
extensive experiments and ablation study to show the advantages of
GC-MVSNet.  
We hope that our work will bring some insights about including explicit geometric reasoning during learning.

\textbf{Acknowledgement:}
This work was supported by Electronics and Telecommunications Research Institute (ETRI) 
grant funded by the Korean government. 
[23ZH1200, The research of the fundamental media·contents technologies for hyper-realistic media space].

\newpage

{\small
\bibliographystyle{ieee_fullname}
\bibliography{paper}

\begin{thebibliography}{10}\itemsep=-1pt

\bibitem{Campbell2008UsingMH}
Neill D.~F. Campbell, George Vogiatzis, Carlos Hern{\'a}ndez, and Roberto
  Cipolla.
\newblock Using multiple hypotheses to improve depth-maps for multi-view
  stereo.
\newblock In {\em European Conference on Computer Vision}, 2008.

\bibitem{chen2019pointbased}
Rui Chen, Songfang Han, Jing Xu, and Hao Su.
\newblock Point-based multi-view stereo network.
\newblock {\em 2019 IEEE/CVF International Conference on Computer Vision
  (ICCV)}, pages 1538--1547, 2019.

\bibitem{Cheng2019USCNet}
Shuo Cheng, Zexiang Xu, Shilin Zhu, Zhuwen Li, Li~Erran Li, Ravi Ramamoorthi,
  and Hao Su.
\newblock Deep stereo using adaptive thin volume representation with
  uncertainty awareness.
\newblock In {\em Proceedings of the IEEE/CVF Conference on Computer Vision and
  Pattern Recognition}, pages 2524--2534, 2020.

\bibitem{Dai2017deformableCNN}
Jifeng Dai, Haozhi Qi, Yuwen Xiong, Yi Li, Guodong Zhang, Han Hu, and Yichen
  Wei.
\newblock Deformable convolutional networks.
\newblock In {\em Proceedings of the IEEE international conference on computer
  vision}, pages 764--773, 2017.

\bibitem{darmon2021wildMVS}
Fran{\c{c}}ois Darmon, B{\'e}n{\'e}dicte Bascle, Jean-Cl{\'e}ment Devaux,
  Pascal Monasse, and Mathieu Aubry.
\newblock Deep multi-view stereo gone wild.
\newblock In {\em 2021 International Conference on 3D Vision (3DV)}, pages
  484--493. IEEE, 2021.

\bibitem{ding2022transmvsnet}
Yikang Ding, Wentao Yuan, Qingtian Zhu, Haotian Zhang, Xiangyue Liu, Yuanjiang
  Wang, and Xiao Liu.
\newblock Transmvsnet: Global context-aware multi-view stereo network with
  transformers.
\newblock In {\em Proceedings of the IEEE/CVF Conference on Computer Vision and
  Pattern Recognition}, pages 8585--8594, 2022.

\bibitem{faugeras1998}
O. Faugeras and R. Keriven.
\newblock Variational principles, surface evolution, pdes, level set methods,
  and the stereo problem.
\newblock {\em IEEE Transactions on Image Processing}, 7(3):336--344, 1998.

\bibitem{fua1995object}
Pascal Fua and Yvan~G Leclerc.
\newblock Object-centered surface reconstruction: Combining multi-image stereo
  and shading.
\newblock {\em International Journal of Computer Vision}, 16(ARTICLE):35--56,
  1995.

\bibitem{Furukawa2010AccurateDA}
Yasutaka Furukawa and Jean Ponce.
\newblock Accurate, dense, and robust multiview stereopsis.
\newblock {\em IEEE Transactions on Pattern Analysis and Machine Intelligence},
  32:1362--1376, 2010.

\bibitem{Galliani2015fusibile}
Silvano Galliani, Katrin Lasinger, and Konrad Schindler.
\newblock Massively parallel multiview stereopsis by surface normal diffusion.
\newblock In {\em 2015 IEEE International Conference on Computer Vision
  (ICCV)}, pages 873--881, 2015.

\bibitem{giang2022curvatureguided}
Khang~Truong Giang, Soohwan Song, and Sungho Jo.
\newblock {CURVATURE}-{GUIDED} {DYNAMIC} {SCALE} {NETWORKS} {FOR}
  {MULTI}-{VIEW} {STEREO}.
\newblock In {\em International Conference on Learning Representations}, 2022.

\bibitem{gu2019casmvsnet}
Xiaodong Gu, Zhiwen Fan, Siyu Zhu, Zuozhuo Dai, Feitong Tan, and Ping Tan.
\newblock Cascade cost volume for high-resolution multi-view stereo and stereo
  matching.
\newblock In {\em Proceedings of the IEEE/CVF conference on computer vision and
  pattern recognition}, pages 2495--2504, 2020.

\bibitem{Hartley2012Multi-view-geometry}
Richard Hartley and Andrew Zisserman.
\newblock {\em Multiple View Geometry in Computer Vision}.
\newblock Cambridge University Press, New York, NY, USA, 2 edition, 2003.

\bibitem{ioffe2015batchnorm}
Sergey Ioffe and Christian Szegedy.
\newblock Batch normalization: Accelerating deep network training by reducing
  internal covariate shift.
\newblock In {\em International conference on machine learning}, pages
  448--456. pmlr, 2015.

\bibitem{jensen2014dtu}
Rasmus Jensen, Anders Dahl, George Vogiatzis, Engil Tola, and Henrik Aan{\ae}s.
\newblock Large scale multi-view stereopsis evaluation.
\newblock In {\em 2014 IEEE Conference on Computer Vision and Pattern
  Recognition}, pages 406--413. IEEE, 2014.

\bibitem{Ji2017surfacenet}
Mengqi Ji, Juergen Gall, Haitian Zheng, Yebin Liu, and Lu Fang.
\newblock Surfacenet: An end-to-end 3d neural network for multiview stereopsis.
\newblock In {\em 2017 IEEE International Conference on Computer Vision
  (ICCV)}, pages 2326--2334, 2017.

\bibitem{kang2001occlusion}
Sing~Bing Kang, R. Szeliski, and Jinxiang Chai.
\newblock Handling occlusions in dense multi-view stereo.
\newblock In {\em Proceedings of the 2001 IEEE Computer Society Conference on
  Computer Vision and Pattern Recognition. CVPR 2001}, volume~1, pages I--I,
  2001.

\bibitem{angelos2020linearattention}
Angelos Katharopoulos, Apoorv Vyas, Nikolaos Pappas, and Fran{\c{c}}ois
  Fleuret.
\newblock Transformers are rnns: Fast autoregressive transformers with linear
  attention.
\newblock In {\em International conference on machine learning}, pages
  5156--5165. PMLR, 2020.

\bibitem{Kingma2014AdamAM}
Diederik~P. Kingma and Jimmy Ba.
\newblock Adam: A method for stochastic optimization.
\newblock {\em CoRR}, abs/1412.6980, 2014.

\bibitem{Knapitsch2017tnt}
Arno Knapitsch, Jaesik Park, Qian-Yi Zhou, and Vladlen Koltun.
\newblock Tanks and temples: Benchmarking large-scale scene reconstruction.
\newblock {\em ACM Transactions on Graphics}, 36(4), 2017.

\bibitem{kutulakos1999spacecarving}
K.N. Kutulakos and S.M. Seitz.
\newblock A theory of shape by space carving.
\newblock In {\em Proceedings of the Seventh IEEE International Conference on
  Computer Vision}, volume~1, pages 307--314 vol.1, 1999.

\bibitem{lhuillier2005quasidense}
M. Lhuillier and L. Quan.
\newblock A quasi-dense approach to surface reconstruction from uncalibrated
  images.
\newblock {\em IEEE Transactions on Pattern Analysis and Machine Intelligence},
  27(3):418--433, 2005.

\bibitem{Lin2016featurepyramid}
Tsung-Yi Lin, Piotr Doll{\'a}r, Ross Girshick, Kaiming He, Bharath Hariharan,
  and Serge Belongie.
\newblock Feature pyramid networks for object detection.
\newblock In {\em Proceedings of the IEEE conference on computer vision and
  pattern recognition}, pages 2117--2125, 2017.

\bibitem{Lin2017FocalLoss}
Tsung-Yi Lin, Priya Goyal, Ross Girshick, Kaiming He, and Piotr Doll{\'a}r.
\newblock Focal loss for dense object detection.
\newblock In {\em Proceedings of the IEEE international conference on computer
  vision}, pages 2980--2988, 2017.

\bibitem{Luo2019Pmvsnet}
Keyang Luo, Tao Guan, Lili Ju, Haipeng Huang, and Yawei Luo.
\newblock P-mvsnet: Learning patch-wise matching confidence aggregation for
  multi-view stereo.
\newblock In {\em 2019 IEEE/CVF International Conference on Computer Vision
  (ICCV)}, pages 10451--10460, 2019.

\bibitem{ma2021eppMVS}
Xinjun Ma, Yue Gong, Qirui Wang, Jingwei Huang, Lei Chen, and Fan Yu.
\newblock Epp-mvsnet: Epipolar-assembling based depth prediction for multi-view
  stereo.
\newblock In {\em 2021 IEEE/CVF International Conference on Computer Vision
  (ICCV)}, pages 5712--5720, 2021.

\bibitem{ma2021epp}
Xinjun Ma, Yue Gong, Qirui Wang, Jingwei Huang, Lei Chen, and Fan Yu.
\newblock Epp-mvsnet: Epipolar-assembling based depth prediction for multi-view
  stereo.
\newblock In {\em Proceedings of the IEEE/CVF International Conference on
  Computer Vision}, pages 5732--5740, 2021.

\bibitem{mi2022gbi}
Zhenxing Mi, Chang Di, and Dan Xu.
\newblock Generalized binary search network for highly-efficient multi-view
  stereo.
\newblock In {\em Proceedings of the IEEE/CVF Conference on Computer Vision and
  Pattern Recognition}, pages 12991--13000, 2022.

\bibitem{nakamura1996stereoocclusion}
Y. Nakamura, T. Matsuura, K. Satoh, and Y. Ohta.
\newblock Occlusion detectable stereo-occlusion patterns in camera matrix.
\newblock In {\em Proceedings CVPR IEEE Computer Society Conference on Computer
  Vision and Pattern Recognition}, pages 371--378, 1996.

\bibitem{peng2022rethinkingMVS}
Rui Peng, Rongjie Wang, Zhenyu Wang, Yawen Lai, and Ronggang Wang.
\newblock Rethinking depth estimation for multi-view stereo: A unified
  representation.
\newblock In {\em Proceedings of the IEEE Conference on Computer Vision and
  Pattern Recognition (CVPR)}, 2022.

\bibitem{qiao2019weightstandardization}
Siyuan Qiao, Huiyu Wang, Chenxi Liu, Wei Shen, and Alan Yuille.
\newblock Weight standardization.
\newblock {\em arXiv preprint arXiv:1903.10520}, 2019.

\bibitem{olaf2015unet}
Olaf Ronneberger, Philipp Fischer, and Thomas Brox.
\newblock U-net: Convolutional networks for biomedical image segmentation.
\newblock In {\em Medical Image Computing and Computer-Assisted
  Intervention--MICCAI 2015: 18th International Conference, Munich, Germany,
  October 5-9, 2015, Proceedings, Part III 18}, pages 234--241. Springer, 2015.

\bibitem{Johannes2016pixelwise}
Johannes~L. Sch{\"o}nberger, Enliang Zheng, Jan-Michael Frahm, and Marc
  Pollefeys.
\newblock Pixelwise view selection for unstructured multi-view stereo.
\newblock In Bastian Leibe, Jiri Matas, Nicu Sebe, and Max Welling, editors,
  {\em Computer Vision -- ECCV 2016}, pages 501--518, Cham, 2016. Springer
  International Publishing.

\bibitem{seitz1997photorealistic}
S.M. Seitz and C.R. Dyer.
\newblock Photorealistic scene reconstruction by voxel coloring.
\newblock In {\em Proceedings of IEEE Computer Society Conference on Computer
  Vision and Pattern Recognition}, pages 1067--1073, 1997.

\bibitem{sinha2007graphcut}
Sudipta~N. Sinha, Philippos Mordohai, and Marc Pollefeys.
\newblock Multi-view stereo via graph cuts on the dual of an adaptive
  tetrahedral mesh.
\newblock In {\em 2007 IEEE 11th International Conference on Computer Vision},
  pages 1--8, 2007.

\bibitem{tola2011largescale}
Engin Tola, Christoph Strecha, and Pascal Fua.
\newblock Efficient large scale multi-view stereo for ultra high resolution
  image sets.
\newblock {\em Machine Vision and Applications}, 23, 09 2011.

\bibitem{tonioni2018realtime}
Alessio Tonioni, Fabio Tosi, Matteo Poggi, Stefano Mattoccia, and Luigi~Di
  Stefano.
\newblock Real-time self-adaptive deep stereo.
\newblock In {\em Proceedings of the IEEE/CVF Conference on Computer Vision and
  Pattern Recognition}, pages 195--204, 2019.

\bibitem{vaswani2017attention}
Ashish Vaswani, Noam Shazeer, Niki Parmar, Jakob Uszkoreit, Llion Jones,
  Aidan~N Gomez, \L~ukasz Kaiser, and Illia Polosukhin.
\newblock Attention is all you need.
\newblock In I. Guyon, U.~Von Luxburg, S. Bengio, H. Wallach, R. Fergus, S.
  Vishwanathan, and R. Garnett, editors, {\em Advances in Neural Information
  Processing Systems}, volume~30. Curran Associates, Inc., 2017.

\bibitem{wang2022mvster}
Xiaofeng Wang, Zheng Zhu, Guan Huang, Fangbo Qin, Yun Ye, Yijia He, Xu Chi, and
  Xingang Wang.
\newblock Mvster: Epipolar transformer for efficient multi-view stereo.
\newblock In {\em European Conference on Computer Vision}, pages 573--591.
  Springer, 2022.

\bibitem{wang2018anytimestereo}
Yan Wang, Zihang Lai, Gao Huang, Brian~H Wang, Laurens Van Der~Maaten, Mark
  Campbell, and Kilian~Q Weinberger.
\newblock Anytime stereo image depth estimation on mobile devices.
\newblock In {\em 2019 international conference on robotics and automation
  (ICRA)}, pages 5893--5900. IEEE, 2019.

\bibitem{wei2021aa}
Zizhuang Wei, Qingtian Zhu, Chen Min, Yisong Chen, and Guoping Wang.
\newblock Aa-rmvsnet: Adaptive aggregation recurrent multi-view stereo network.
\newblock In {\em Proceedings of the IEEE/CVF International Conference on
  Computer Vision}, pages 6187--6196, 2021.

\bibitem{weilharter2021highresMVS}
Rafael Weilharter and Friedrich Fraundorfer.
\newblock Highres-mvsnet: A fast multi-view stereo network for dense 3d
  reconstruction from high-resolution images.
\newblock {\em IEEE Access}, 9:11306--11315, 2021.

\bibitem{he2018groupnorm}
Yuxin Wu and Kaiming He.
\newblock Group normalization.
\newblock In {\em Proceedings of the European conference on computer vision
  (ECCV)}, pages 3--19, 2018.

\bibitem{xu2021nonlocalrecurrent}
Qingshan Xu, Martin~R. Oswald, Wenbing Tao, Marc Pollefeys, and Zhaopeng Cui.
\newblock Non-local recurrent regularization networks for multi-view stereo.
\newblock {\em CoRR}, abs/2110.06436, 2021.

\bibitem{xu2019multiscale}
Qingshan Xu and Wenbing Tao.
\newblock Multi-scale geometric consistency guided multi-view stereo.
\newblock In {\em Proceedings of the IEEE/CVF Conference on Computer Vision and
  Pattern Recognition}, pages 5483--5492, 2019.

\bibitem{yan2020dynamicfusion}
Jianfeng Yan, Zizhuang Wei, Hongwei Yi, Mingyu Ding, Runze Zhang, Yisong Chen,
  Guoping Wang, and Yu-Wing Tai.
\newblock Dense hybrid recurrent multi-view stereo net with dynamic consistency
  checking.
\newblock In {\em European conference on computer vision}, pages 674--689.
  Springer, 2020.

\bibitem{yang2019CVPMVS}
Jiayu Yang, Wei Mao, Jose~M. Alvarez, and Miaomiao Liu.
\newblock Cost volume pyramid based depth inference for multi-view stereo.
\newblock In {\em The IEEE/CVF Conference on Computer Vision and Pattern
  Recognition (CVPR)}, June 2020.

\bibitem{yao2018mvsnet}
Yao Yao, Zixin Luo, Shiwei Li, Tian Fang, and Long Quan.
\newblock Mvsnet: Depth inference for unstructured multi-view stereo.
\newblock In {\em Proceedings of the European Conference on Computer Vision
  (ECCV)}, pages 767--783, 2018.

\bibitem{yao2019recurrent}
Yao Yao, Zixin Luo, Shiwei Li, Tianwei Shen, Tian Fang, and Long Quan.
\newblock Recurrent mvsnet for high-resolution multi-view stereo depth
  inference.
\newblock {\em Computer Vision and Pattern Recognition (CVPR)}, 2019.

\bibitem{yao2019blended}
Yao Yao, Zixin Luo, Shiwei Li, Jingyang Zhang, Yufan Ren, Lei Zhou, Tian Fang,
  and Long Quan.
\newblock Blendedmvs: A large-scale dataset for generalized multi-view stereo
  networks.
\newblock In {\em Proceedings of the IEEE/CVF conference on computer vision and
  pattern recognition}, pages 1790--1799, 2020.

\bibitem{yin2018hierarchical}
Zhichao Yin, Trevor Darrell, and Fisher Yu.
\newblock Hierarchical discrete distribution decomposition for match density
  estimation.
\newblock In {\em Proceedings of the IEEE/CVF conference on computer vision and
  pattern recognition}, pages 6044--6053, 2019.

\bibitem{YU2021AAcostvolume}
Anzhu Yu, Wenyue Guo, Bing Liu, Xin Chen, Xin Wang, Xuefeng Cao, and Bingchuan
  Jiang.
\newblock Attention aware cost volume pyramid based multi-view stereo network
  for 3d reconstruction.
\newblock {\em ISPRS Journal of Photogrammetry and Remote Sensing},
  175:448--460, 2021.

\bibitem{zhang2008consistentvideo}
Guofeng Zhang, Jiaya Jia, Tien-Tsin Wong, and Hujun Bao.
\newblock Recovering consistent video depth maps via bundle optimization.
\newblock In {\em 2008 IEEE Conference on Computer Vision and Pattern
  Recognition}, pages 1--8, 2008.

\bibitem{Zhang2020visibility}
Jingyang Zhang, Yao Yao, Shiwei Li, Zixin Luo, and Tian Fang.
\newblock Visibility-aware multi-view stereo network.
\newblock {\em British Machine Vision Conference (BMVC)}, 2020.

\bibitem{zhu2018deformableV2}
Xizhou Zhu, Han Hu, Stephen Lin, and Jifeng Dai.
\newblock Deformable convnets v2: More deformable, better results.
\newblock In {\em Proceedings of the IEEE/CVF conference on computer vision and
  pattern recognition}, pages 9308--9316, 2019.

\end{thebibliography}
}

\end{document}


\title{ Multi-View, Multi-Scale, Geometrically-Consistent
  Multi-View Stereo (Supplementary Materials)} 

  \author{Vibhas K. Vats, Sripad Joshi, David J. Crandall\\
Indiana University Bloomington\\
{\tt\small \{vkvats, joshisri, djcran\}@iu.edu}
\and
Md. Alimoor Reza\\
Drake University\\
{\tt\small md.reza@drake.edu}
\and 
Soon-heung Jung\\
ETRI\\
{\tt\small zeroone@etri.re.kr}
}

\newcommand\myfigure{%
  \includegraphics[width=1\textwidth,height=15\baselineskip]{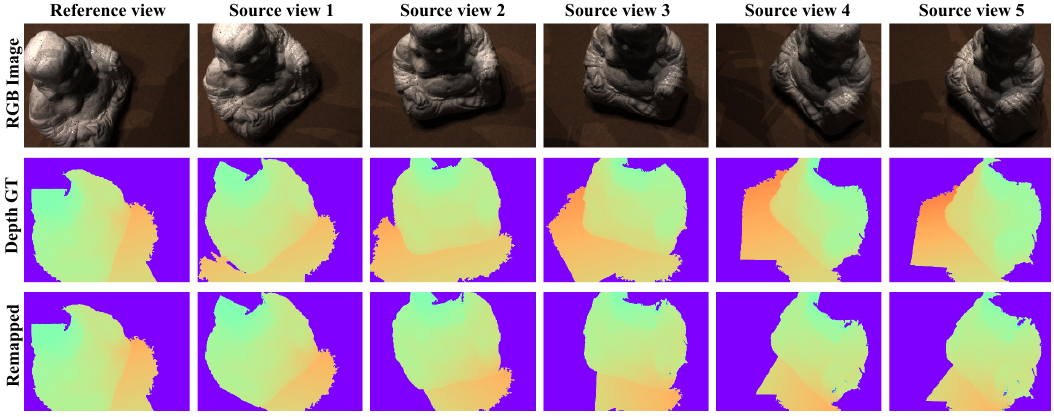}\\
  \refstepcounter{figure}\normalfont Figure~\thefigure: First row shows the selection of $M$ closest source images for a given reference image. Middle row shows the corresponding ground truth depth maps and last row shows the remapped source ground truth depth maps using x-y coordinates of reference view projection to the source view. During remapping, all additional pixels from the source views are ignored. The remapped depths are then back-projected to source view to generate mask. Finally, reference view mask is applied on per-pixel penalty to restrict the penalties. Corresponding final $\xi_p$ is shown in Fig. 3 of the paper. All depth maps are shown within respective view mask.
  \label{fig:occlusion-handling}\\
}

\makeatletter
\let\@oldmaketitle\@maketitle
\renewcommand{\@maketitle}{\@oldmaketitle
    \myfigure}
\makeatother

\maketitle
\thispagestyle{empty}
\appendix

\section{Occlusion and its impact}

Modeling occlusion of pixels in multi-view setting is a difficult problem. It is difficult to reason about a pixel in a view whose corresponding 3D points are occluded in other view. The problem becomes significant if a penalty is being attached to all such pixels, like in the proposed multi-view geometric consistency checking module. The GC module checks geometric consistency of each pixel across multiple source views and awards a penalty for inconsistency. Assigning penalties to occluded pixels and multiplying it with depth error adversely impacts the training process. Early in our experiments, we observe that the loss started to explode with training, i.e. as the model trains the loss values starts to increase. 

Our investigation suggests that the wrongful penalties of occluded pixels dominated loss during training. We find that our method becomes robust to this problem with a series of steps taken. First, we use the closest source view images as defined by MVSNet \cite{yao2018mvsnet}. The first row of Fig. \ref{fig:occlusion-handling} shows the source view selection for the given reference view. Choosing closest view to the reference view reduces the number of possible occluded pixels. Second, during forward-backward-reprojection, we remap the source view depth map as per the x-y coordinate projections of the reference view to the source view and then, the remapped values are back-projected to the reference view (see Alg. 2 in the paper). The last row in Fig. \ref{fig:occlusion-handling} shows the remapped version of the source view depth maps. During remapping, all the occluded as well as the additional pixels of the source view is dropped and then this remapped version is back-projected. This handles the extreme cases of occlusion or additional visible pixels. At the end, once the per-pixel penalty is generated, we apply the reference view binary mask on it to do away with any such pixel which is not part of the scene in consideration (see Fig. 3 in the paper). The combination of these steps help us control the impact of wrongful penalties and stabilize the training process.

\section{Geometric Consistency Module}

\begin{figure}[ht]
\begin{center}
   \includegraphics[width=0.85\linewidth]{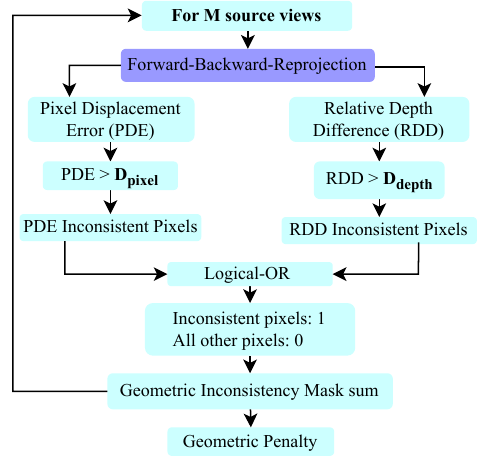}
\end{center}
    \vspace{-15pt}
   \caption{GC module flow-chart for consistency check.}
\label{fig:gc-module-flow-chart}
\vspace{-10pt}
\end{figure}

We describe the steps of geometric consistency (GC) module in Fig. \ref{fig:gc-module-flow-chart}. At each stage, the geometric consistency of estimated depth map is checked across $M$ source views. For each source view, we perform the forward-backward-reprojection of estimated depth map to reason about geometric inconsistency of pixels (described in Alg. 2). In this three-step process, first, we warp each pixel $P_{0}$ of a reference view depth map $D_{0}$ to its $i^{th}$ neighboring source view to obtain corresponding pixel $P^{'}_{i}$. Then, we back-project $P^{'}_{i}$ into 3D space and finally, we reproject it to the reference view as $P^{"}_{0}$ using $c_0$. $D_0$, $D^{'}_{P^{'}_{i}}$ and $D^{"}_{P^{"}_{0}}$ represents depth value of pixels associated with $P_0$, $P^{'}_{i}$ and $P^{"}_{0}$ \cite{Hartley2012Multi-view-geometry}. With $P^{"}_{0}$ and $D^{"}_{P^{"}_{0}}$, we calculate pixel displacement error (PDE) and relative depth difference (RDD). After taking logical-OR between PDE and RDD, we assign value $1$ to all inconsistent pixel and zero to all other pixels. The geometric inconsistency mask sum is generated over $M$ source views and  averaged to generate per-pixel penalty $\xi_p$.

\section{Depth Interval Ratio (DIR)}

\begin{table}[ht]
  \begin{center}
    {\small{
\begin{tabular}{clccc}
\toprule
$\xi_p$ Range  & Stage-wise DIR  & Acc$\downarrow$ & Comp$\downarrow$ & Overall$\downarrow$ \\
\midrule
$[1,3]$ & 2.0, 0.8, 0.40 & 0.338 & 0.269 & 0.3035 \\
$[1,3]$ & 2.0, 0.7, 0.35 & 0.343 & \textbf{0.264} & 0.3035 \\
$[1,3]$ & 2.0, 0.7, 0.30 & 0.331 & 0.27 & 0.3005 \\ 
$[1,3]$ & 1.6, 0.7, 0.30 & \textbf{0.329} & 0.271 & \textbf{0.300} \\
\bottomrule
\end{tabular}}}
\vspace{-5pt}
\caption{The performance of GC-MVSNet on evaluation set of DTU \cite{jensen2014dtu} with change in stage-wise DIR (depth interval ratio).}
\label{table:dir}
\vspace{-22pt}
\end{center}
\end{table}

DIR directly impacts the separation of two hypothesis planes at pixel level. For a given stage, the pixel-level depth interval is calculated as product of $DIR_{stage}$ and \textit{depth interval} (DI). The value of DI is calculated using \textit{interval scale} and a constant value provided in DTU camera parameter files. 

Following the trend of modern learning-based methods \cite{xu2019multiscale, gu2019casmvsnet, yao2018mvsnet, YU2021AAcostvolume, ding2022transmvsnet, peng2022rethinkingMVS, Cheng2019USCNet}, we train our model on $512 \times 640$ resolution and test on $864 \times 1152$ resolution.  To adjust for the pixel-level depth interval caused by the increase in resolution, we explore different DIR values for testing on DTU. We train our model with stage-wise DIR $2.0, 0.8, 0.4$ ($DIR_{train}$).  such that the refine stage pixel-level depth interval is same as the provided \textit{interval scale} value of $1.06$. Table \ref{table:dir} shows DIR values for evaluation on DTU, we only explore smaller values than $DIR_{train}$ to compensate for the increase in resolution. GC-MVSNet achieves its optimal performance at DIR $1.6, 0.7, 0.3$ with $\xi_p \in [1,3]$, $DIR_{test}$. We use the same $DIR_{train}$ and $DIR_{test}$ with $\xi_p \in [1,2]$.

\section{Stabilizing the Training Process}

\begin{figure}[ht]
\begin{center}
   \includegraphics[width=1\linewidth]{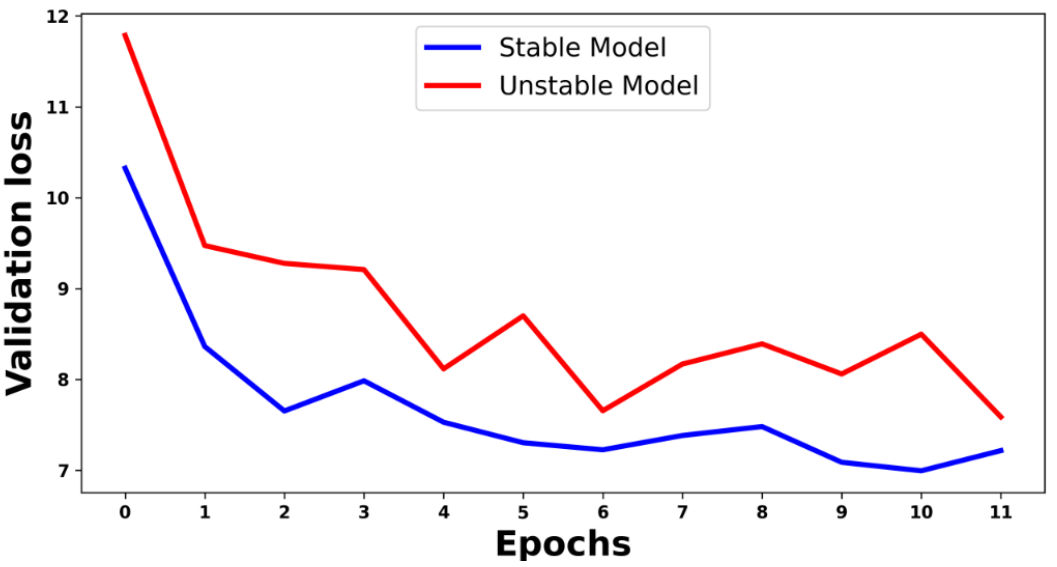}
\end{center}
    \vspace{-18pt}
   \caption{Validation loss on DTU \cite{jensen2014dtu} dataset during training. The red line shows the unstable model training, validation loss change in zig-zag manner. Blue line shows stable training with smooth change in validation loss.}
\label{fig:stable-training}
\vspace{-10pt}
\end{figure}

Most of the modern learning-based MVS methods \cite{gu2019casmvsnet, ding2022transmvsnet, peng2022rethinkingMVS, wei2021aa, Zhang2020visibility, weilharter2021highresMVS} use BatchNorm \cite{ioffe2015batchnorm} along with Apex (Nvidia) for batch synchronization. BatchNorm is most useful with large batch size. For smaller batch size, like $1$ or $3$, it degrades the training process \cite{ioffe2015batchnorm} by poor estimation of population mean ($\mu$) and std. ($\sigma$) over small batch size. 

GroupNorm \cite{he2018groupnorm} alleviates this problem by estimating $\mu$ and $\sigma$ along the channels instead of batch. Weight-standardization \cite{qiao2019weightstandardization} further stabilizes the training and evaluation steps. We refer to the original papers for further understanding of these concepts. GC-MVSNet replaces BatchNorm with GroupNorm and Weight-standardization techniques to stabilize the training process. Fig. \ref{fig:stable-training} shows the difference between model trained with (red line) and without (blue line) BatchNorm. With the use of GroupNorm along with Weight-standardization, the evaluation loss curve become smooth and stable.

\section{Depth Map Fusion Methods}

The quality of point clouds depends heavily on depth fusion methods and their hyperparameters. Following the recent learning-based methods \cite{ding2022transmvsnet, peng2022rethinkingMVS, gu2019casmvsnet}, we also use different fusion method for DTU and Tanks and Temples dataset. For DTU, we use Fusibile \cite{Galliani2015fusibile} and for Tanks and Temples, we use Dynamic method \cite{ding2022transmvsnet, wei2021aa}.

Fusibile fusion method uses three hyperparameters, disparity threshold, probability confidence threshold, and consistency threshold. Disparity threshold defines the upper limit of disparity for points to be eligible for fusion. Probability confidence threshold defines the lower limit of confidence above which points are eligible for fusion. The consistency threshold mandates that the eligible points be geometrically consistent across as many source views. During the fusion process, only those points that satisfy all three conditions are fused into point cloud. 

Dynamic fusion method uses only two hyperparameters, probability confidence threshold and consistency threshold. Both these hyperparameters have exact same function as in Fusibile method. The disparity threshold is not provided by the user, it is dynamically adjusted during the fusion process. 

\section{Accuracy and Completeness Metrics}

\begin{figure}[ht]
\begin{center}
   \includegraphics[width=0.9\linewidth]{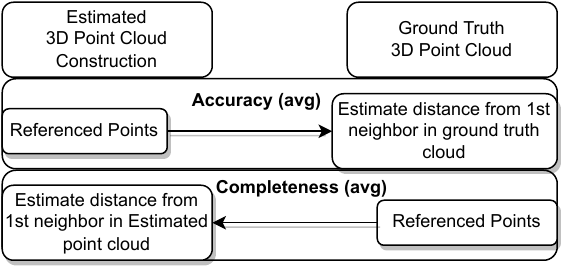}
\end{center}
    \vspace{-18pt}
   \caption{The process of calculating accuracy and completeness for DTU \cite{jensen2014dtu} point cloud evaluation.}
\label{fig:acc-comp-diagram}
\vspace{-10pt}
\end{figure}

Accuracy and completeness are two metrics used with DTU \cite{jensen2014dtu} dataset. Fig. \ref{fig:acc-comp-diagram} shows the process of calculation. Accuracy is the average of the distance of the first neighbor from predicted point cloud to ground truth point cloud. It only considers the points which are below the maximum threshold for the distance. For completeness, same process is repeated but with ground truth as referenced point cloud, i.e. the average of the distance of the first neighbor from the ground truth point cloud to the predicted point cloud.

\section{Use of Existing Assets}

We use PyTorch to implement GC-MVSNet. It is based on CasMVSNet \cite{gu2019casmvsnet} and TransMVSNet \cite{ding2022transmvsnet}. These two methods heavily borrow code from the PyTorch implementation of MVSNet \cite{yao2018mvsnet}. 

We use preprocessed images and camera parameters of DTU \cite{jensen2014dtu} dataset from official repository of MVSNet \cite{yao2018mvsnet} and R-MVSNet \cite{yao2019recurrent}. We follow \cite{darmon2021wildMVS} for training and testing on BlendedMVS \cite{yao2019blended}. For Tanks and Temples \cite{Knapitsch2017tnt} evaluation, we use images and camera parameters as used in R-MVSNet \cite{yao2019recurrent}.

\section{Point Clouds}

\begin{figure*}[t]
\begin{center}
    \includegraphics[width=0.85\textwidth]{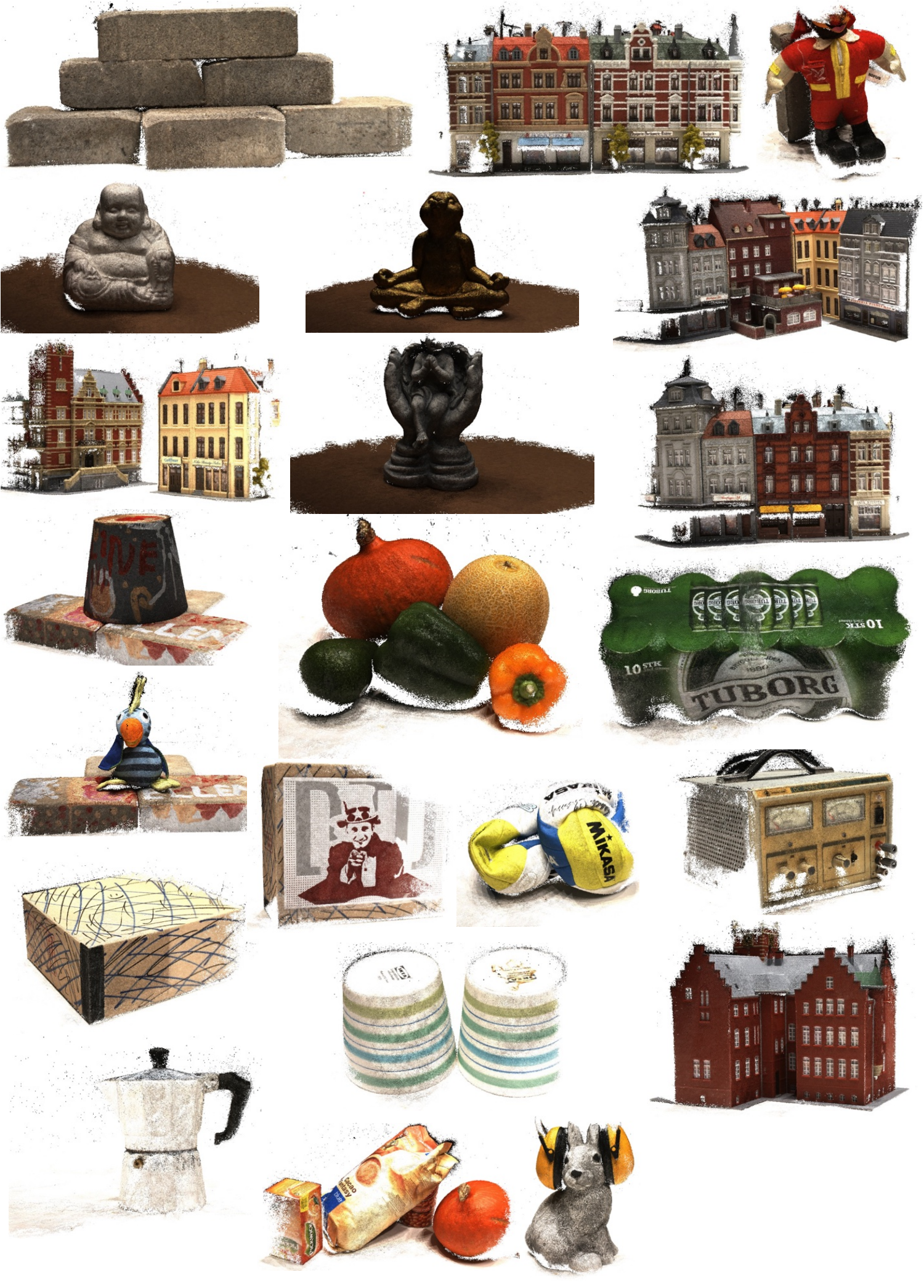}
    \caption{Point clouds reconstructed using GC-MVSNet for all scenes from DTU \cite{jensen2014dtu} evaluation set.}
    \label{fig:dtu-point-clouds}
\end{center}
\end{figure*}

\begin{figure*}[t]
\begin{center}
    \includegraphics[width=0.98\textwidth]{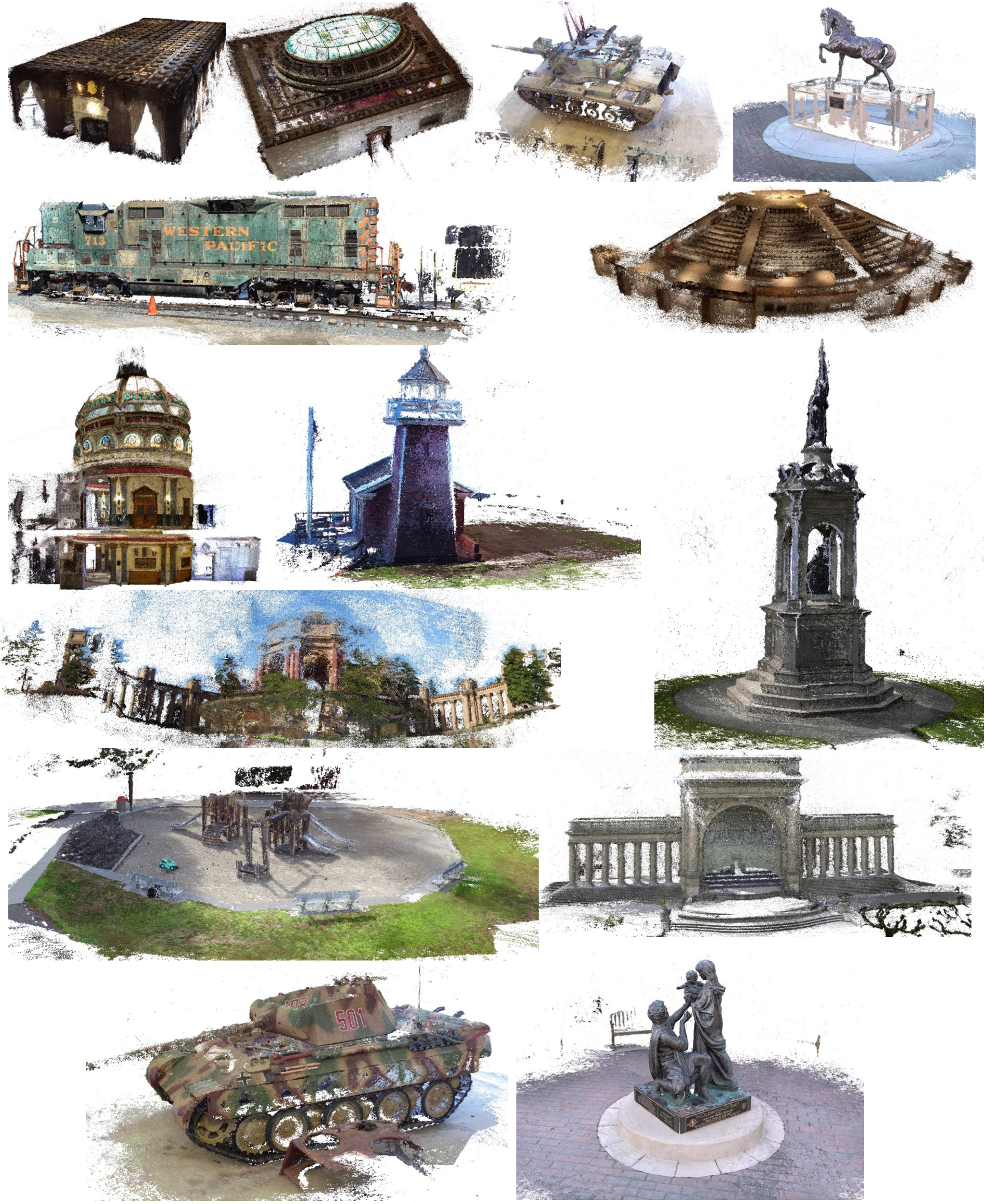}
    \caption{Point clouds reconstructed using GC-MVSNet for all scenes from Tanks and Temples \cite{Knapitsch2017tnt} intermediate and advanced set.}
    \label{fig:tnt-point-clouds}
\end{center}
\end{figure*}

\begin{figure*}[!ht]
\begin{center}
    \includegraphics[width=0.95\textwidth]{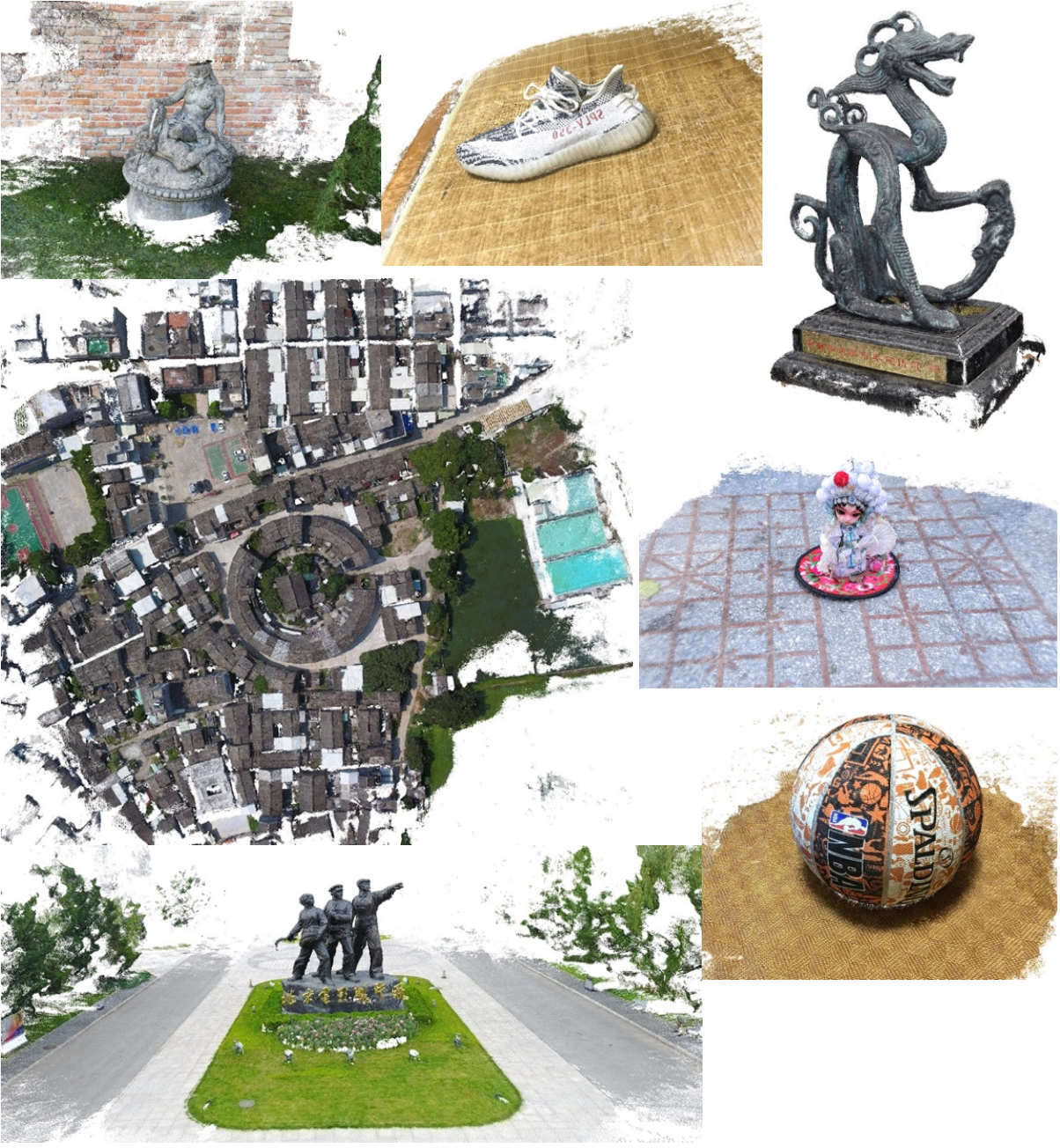}
    \caption{Point clouds reconstructed using GC-MVSNet for all scenes from BlendedMVS \cite{yao2019blended} evaluation set.}
    \label{fig:blended-point-clouds}
\end{center}
\end{figure*}

In this section, we show all evaluation set points clouds reconstructed using GC-MVSNet on DTU \cite{jensen2014dtu}, Tanks and Temples \cite{Knapitsch2017tnt} and BlendedMVS \cite{yao2019blended} datasets. Fig. \ref{fig:dtu-point-clouds}, \ref{fig:tnt-point-clouds} and \ref{fig:blended-point-clouds} show all evaluation set point clouds from DTU, Tanks and Temples and BlendedMVS, respectively.

\newpage

{\small
\bibliographystyle{ieee_fullname}
\bibliography{supplementary}
}